\def\year{2017}
\documentclass[letterpaper]{article}
\usepackage{aaai17}
\usepackage{times}
\usepackage{helvet}
\usepackage{courier}
\usepackage{graphicx}
\usepackage{amssymb}
\usepackage{amsmath}
\usepackage[linesnumbered]{algorithm2e}
\usepackage{mdwlist}
\usepackage{url}
\usepackage{ wasysym }
\usepackage{setspace}
\usepackage{color}
\usepackage{textcomp}
\usepackage{booktabs}
\usepackage{ccicons, xcolor, soul}
\usepackage{wrapfig}
\usepackage{subcaption}
\usepackage{sidecap}
\usepackage{multirow}
\usepackage{hhline}

\makeatletter
\newcommand{\xRightarrow}[2][]{\ext@arrow 0359\Rightarrowfill@{#1}{#2}}
\makeatother

\begin{document}
\title{Tweeting AI: Perceptions of AI-Tweeters (AIT) vs Expert AI-Tweeters (EAIT)}
\author{ Lydia Manikonda, Cameron Dudley, Subbarao Kambhampati \\ Arizona State University, Tempe, AZ \\ \{lmanikon, cjdudley, rao\}@asu.edu} 

\nocopyright

\maketitle
\begin{abstract}
\begin{quote}
With the recent advancements in Artificial Intelligence (AI), various organizations and individuals started debating about the progress of AI as a blessing or a curse for the future of the society. This paper conducts an investigation on how the public perceives the progress of AI by utilizing the data shared on Twitter. Specifically, this paper performs a comparative analysis on the understanding of users from two categories -- general AI-Tweeters (AIT) and the expert AI-Tweeters (EAIT) who share posts about AI on Twitter. Our analysis revealed that users from both the categories express distinct emotions and interests towards AI. Users from both the categories regard AI as positive and are optimistic about the progress of AI but the experts are more negative than the general AI-Tweeters. Characterization of users manifested that `London' is the popular location of users from where they tweet about AI. Tweets posted by AIT are highly retweeted than posts made by EAIT that reveals greater diffusion of information from AIT.
\end{quote}
\end{abstract}

\section{Introduction}
Due to the rapid progress of technology especially in the field of AI, there are various discussions and concerns about the threats and benefits of AI. Some of these discussions include -- improving the everyday lives of individuals (https://goo.gl/ViLdgV), ethical issues associated with the intelligent systems (https://goo.gl/5KlmXk), etc. Social media platforms are ideal repositories of opinions and discussion threads. Twitter is one of the popular platforms  where individuals post their statuses, opinions and perceptions about the ongoing issues in the society~\cite{Java2007,Naaman2010,yang2010predicting}. This paper focuses on investigating the Twitter posts on AI to understand the perceptions of individuals. Specifically, we compare and contrast the perceptions of users from two categories -- general AI-Tweeters (AIT) and expert AI-Tweeters (EAIT). We believe that the findings from this analysis can help research funding agencies, organizations, industries who are curious about the public perceptions of AI. 

Efforts to understand public perceptions of AI are not new. Recent work by Fast et. al~\cite{EthanAAAI2016} conducts a longitudinal study about articles published on AI from New York Times between January 1986 and May 2016 are transformed over the years. This study revealed that from 2009 the discussion on AI has sharply increased and is more optimistic than pessimistic. It also disclosed that fears about losing control over AI systems have been increasing in the recent years. Another recent survey~\cite{LeslieHarvard2016} conducted by the Harvard Business Review on individuals who do not have any background in technology, stated the positive perceptions towards AI. In contrast, even though online social media platforms are the main channels of communication~\cite{Java2007,Naaman2010,yang2010predicting}, there is no existing work on how and what users share about AI on these platforms. This paper attempts to learn the perceptions of individuals manifested by their posts shared on Twitter. 

Towards this goal, we attempt to answer 5 important questions through a thorough quantitative and comparative investigation of the posts shared on Twitter. 1) What are the insights that could be learned by characterizing the individuals and their interests who are making AI-related posts? 2) What is the Twitter engagement rate for the AI-tweets? 3) Are the posts about AI optimistic or pessimistic? 4) What are the most interesting topics of discussion to the users? 5) What can we learn about the frequently co-occurring terms with AI vocabulary? We address each of these questions in the next few sections.

Our analysis reveals intriguing differences between the posts shared by AIT and EAIT on Twitter. Specifically, this analysis reveals four interesting findings about the perceptions of individuals who are using Twitter to share their opinions about AI. Firstly, users from both the categories are emotionally positive (or optimistic) towards the progress of AI. Secondly, even though users are positive overall, users from EAIT are more negative than those from AIT. Thirdly, we found that the tweets shared by EAIT have lower diffusion rate than the tweets posted by AIT as measured by the magnitude of retweets. Fourth, and last, users from AIT are geographically distributed (mostly Europe and US) with London and New York City taking the top positions.

In the next section, we describe the process of collecting data from the two categories of users that we use for the study presented in this paper. We compare and contrast each of the 5 research questions we posed for Twitter users from the two categories. Results of a similar analysis conducted on the AI posts shared on Reddit, is attached as an appendix. 



\section{Data Collection}

\subsection{AI-Tweeters (AIT)}
We employ the official API of Twitter~\footnote{https://dev.twitter.com/overview/api} along with a frequency-based hashtag selection approach to crawl the data from common Twitter users who tweet about AI. We first identify an appropriate set of hashtags that focus on the artificial intelligence in social media. In our case, these are -- \#ai and \#artificialintelligence. With these seed hashtags, we crawled 2 million unique tweets. Using the hashtags presented in these tweets, we iteratively obtained the co-occurring hashtags and sort them based on their frequency. We remove non-technical hashtags included in this sorted hashtag list for example: \#trump, \#politics, etc. The top-15 co-occurring hashtags after the pre-processing are shown in Table~\ref{tab:cooccurringhts}. 

\begin{table}[h]
\small
\centering
\setlength{\tabcolsep}{3pt}
\begin{tabular}{l l l} \hline
1. \#ai & 2. \#artificialintelligence & 3. \#machinelearning \\
4. \#bigdata & 5. \#iot & 6. \#deeplearning \\
7. \#robotics & 8. \#datascience & 9. \#cybersecurity \\
10. \#vr & 11. \#ar & 12. \#nlp \\
13. \#ux & 14. \#algorithms  & 15. \#socialmedia \\ \hline
\end{tabular}
\caption{Top-15 co-occurring hashtags with the seed hashtags: \#ai and \#artificialintelligence}
\label{tab:cooccurringhts}
\end{table}

We used the top-4 hashtags from this list: \emph{\#ai}, \emph{\#artificialintelligence}, \emph{\#machinelearning} and \emph{\#bigdata} as the final hashtag set to crawl a set of 2.3 million tweets. In our dataset we found that the set of tweets obtained using these 4 hashtags are the super set of all the tweets crawled by utilizing the remaining hashtags presented in Table~\ref{tab:cooccurringhts}. Each tweet in this dataset is public and contains the following post-related information: 

\begin{itemize}
\small
\item tweet id 
\vspace{-0.035in}
\item posting date
\vspace{-0.035in}
\item number of favorites received
\vspace{-0.035in}
\item number of times it is retweeted
\vspace{-0.035in}
\item the url links shared as a part of it
\vspace{-0.035in}
\item text of the tweet including the hashtags
\vspace{-0.035in}
\item geolocation if tagged
\end{itemize} 

A tweet may contain more than a single hashtag. From this set of tweets, we remove the redundant tweets that have more than one of these four hashtags we considered. This resulted in a dataset of 0.2 million tweets that are unique and are posted by a unique set of 33K users. Due to the download limit of the Twitter API, all the tweets in our dataset are from February 2017, and none of the tweets we crawled were tagged with a geolocation.

\begin{table}[h]
\centering
\small
\begin{tabular}{ l | c | c }  
\textbf{Metric} & \textbf{AIT} & \textbf{EAIT} \textbf{} \\ \hline 
\textit{Mean} & 10.9 (67.1) & 13.96 (72.9) \\ \hline 
\textit{Median} & 14 (89) & 15.0 (77.0) \\ \hline 
\textit{Min.} & 1 (1) & 1 (1) \\ \hline 
\textit{Max.} & 28 (129) & 46 (126) \\ \hline 
\end{tabular}
\caption{Statistics about the number of words in a post (and number of characters in a post) crawled from Twitter for AIT vs EAIT}
\label{tab:dataStats}
\end{table}

\subsection{Expert AI-Tweeters (EAIT)}
The team at IBM Watson has recently shared an article (https://goo.gl/PdRlHT) listing 30 people in AI as ``AI Influencers 2017''. This article suggested that these are the top-30 people in AI to follow on Twitter to get updated information about AI. For the purposes of our investigation, we contacted the author of this article through a private email to learn the approach used to compile this list. According to the author, this list was compiled by using a combination of thought leaders that the author's team was aware of. This list includes -- AI experts who are most active on Twitter, influencers who speak regularly and present research at AI events \& conferences, and entrepreneurs in the AI space who are working on interesting products and technology. The IBM team also interviewed multiple experts to ask them who they follow on Twitter to stay updated on the trending news about AI. Table~\ref{tab:top30ai} enlists the AI Influencers 2017 named by this article (sorted alphabetically). We then utilize the Twitter official API to crawl the timelines of these 30 AI influencers. 

\begin{table}[h]
\sffamily
\scriptsize
\centering
\setlength{\tabcolsep}{3pt}
\begin{tabular}{l l l} \hline
Nathan Benaich & Joanna Bryson & Alex Champandard \\
Soumith Chintala & Adam Coates & The CyberCode Twins \\
Jana Eggers & Oren Etzioni & Martin Ford \\ 
Adam Geitgey & John C. Havens & J.J Kardwell \\
David Kenny & Tessa Lau & Dr Angelica Lim \\
Satya Mallick & Gary Marcus & Chris Messina \\
Elon Musk & Dr Andy Pardoe & Alec Radford \\
Delip Rao & Dr Roman Yampolskiy & Gideon Rosenblatt \\
Adam Rutherford & Matt Schlicht & Murray Shanahan \\
Amir Shevat & Rest Sidhu & Adelyn Zhou \\ \hline
\end{tabular}
\caption{Top-30 AI Influencers 2017 enlisted by IBM}
\label{tab:top30ai}
\end{table}


\begin{figure*}[!ht]
\centering
\includegraphics[height=3.6in]{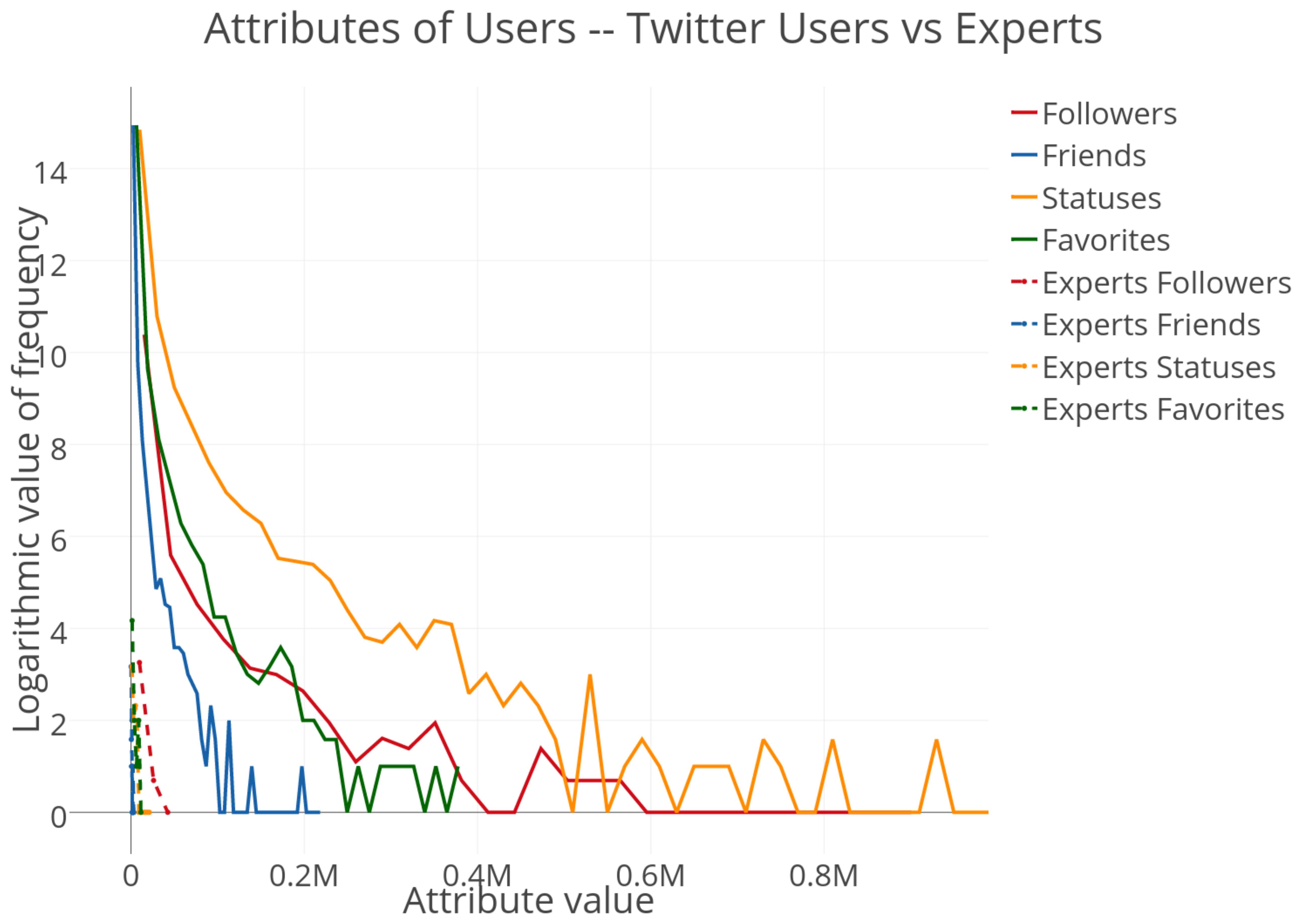}
\caption{User attributes -- Followers; Friends; Statuses; Favorites. The plain lines correspond to the Twitter users and the dotted lines correspond to the experts. X-axis represents the attribute's value where as, Y-axis represents the logarithmic value of the frequency}
\label{fig:userattr}
\end{figure*}

\section{RQ1: Characterization of Users}
Before we delve deep in to investigating the research questions posed earlier, we present few details about the demographics of users from both the categories. We first focus on the influence attributes of users -- \#statuses shared, \#followers, \#friends, \#favorites. To understand the differences between the two types of user categories based on their activity and influence of tweets, we plot the logarithmic frequencies of these attributes in Figure~\ref{fig:userattr}. Surprisingly, the influence attribute values of experts are significantly lower than the general Twitter users.

Figure~\ref{fig:userattr} shows that both sets of users are highly active on Twitter by sharing statuses and favoriting tweets. When we consider the other influence attributes -- followers and friends, on an average both sets of users have large number of followers than friends (users you are following). The statuses shared by EAIT are approximately equal to the number of favorites. However, AIT share more number of statuses than favoriting other tweets. This observation also suggests that the users in AIT and EAIT are active on Twitter as user activity is influential in attracting more number of followers~\cite{Bakshy2011EIQ}.
 
\begin{figure}[!h]
        \centering
    \begin{subfigure}[t]{\columnwidth}
        \centering
        \includegraphics[height=1.7in]{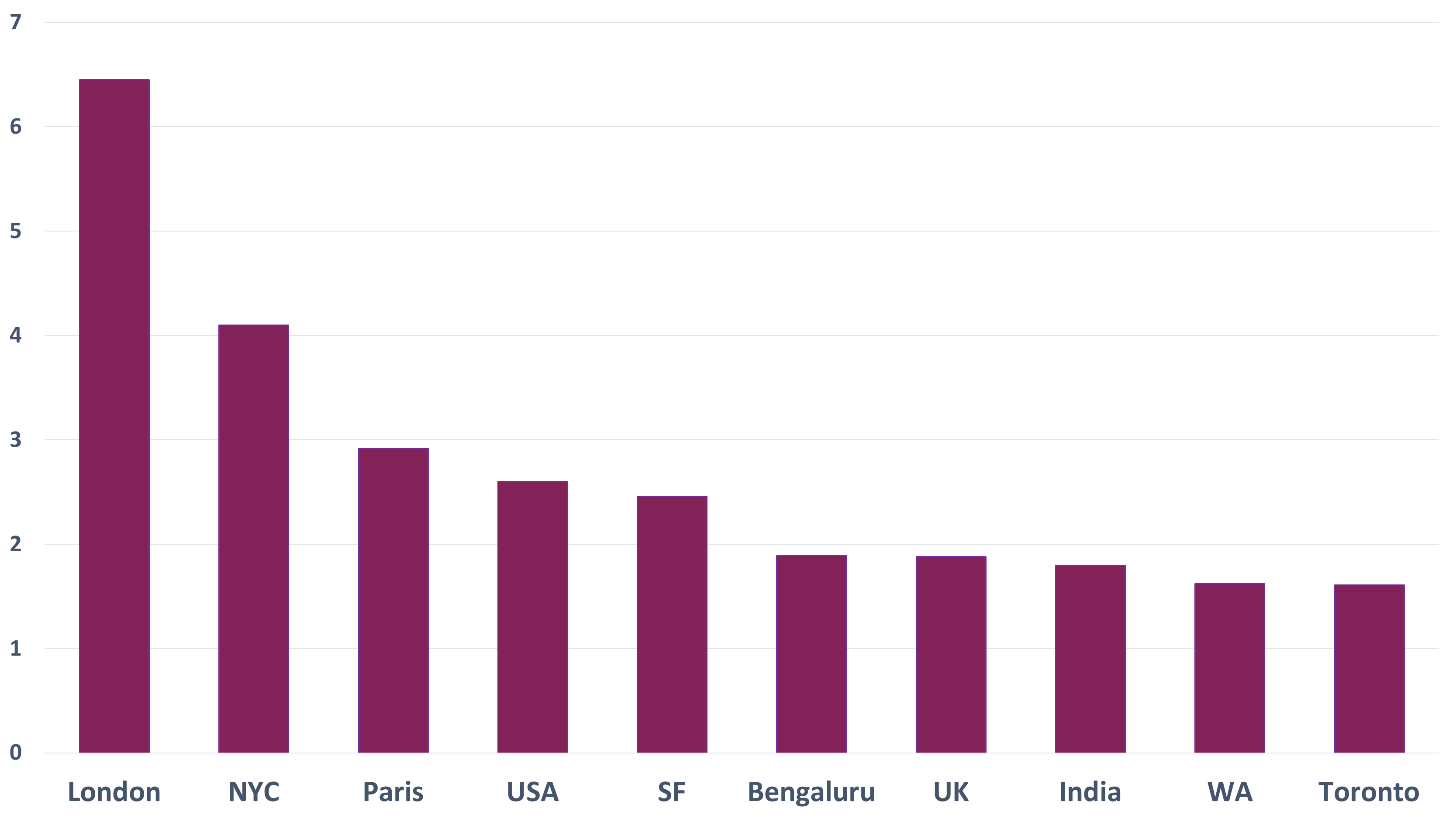}
        \caption{Twitter Users}
    \end{subfigure} \\
    ~
    \begin{subfigure}[t]{\columnwidth}
        \centering
        \includegraphics[height=1.7in]{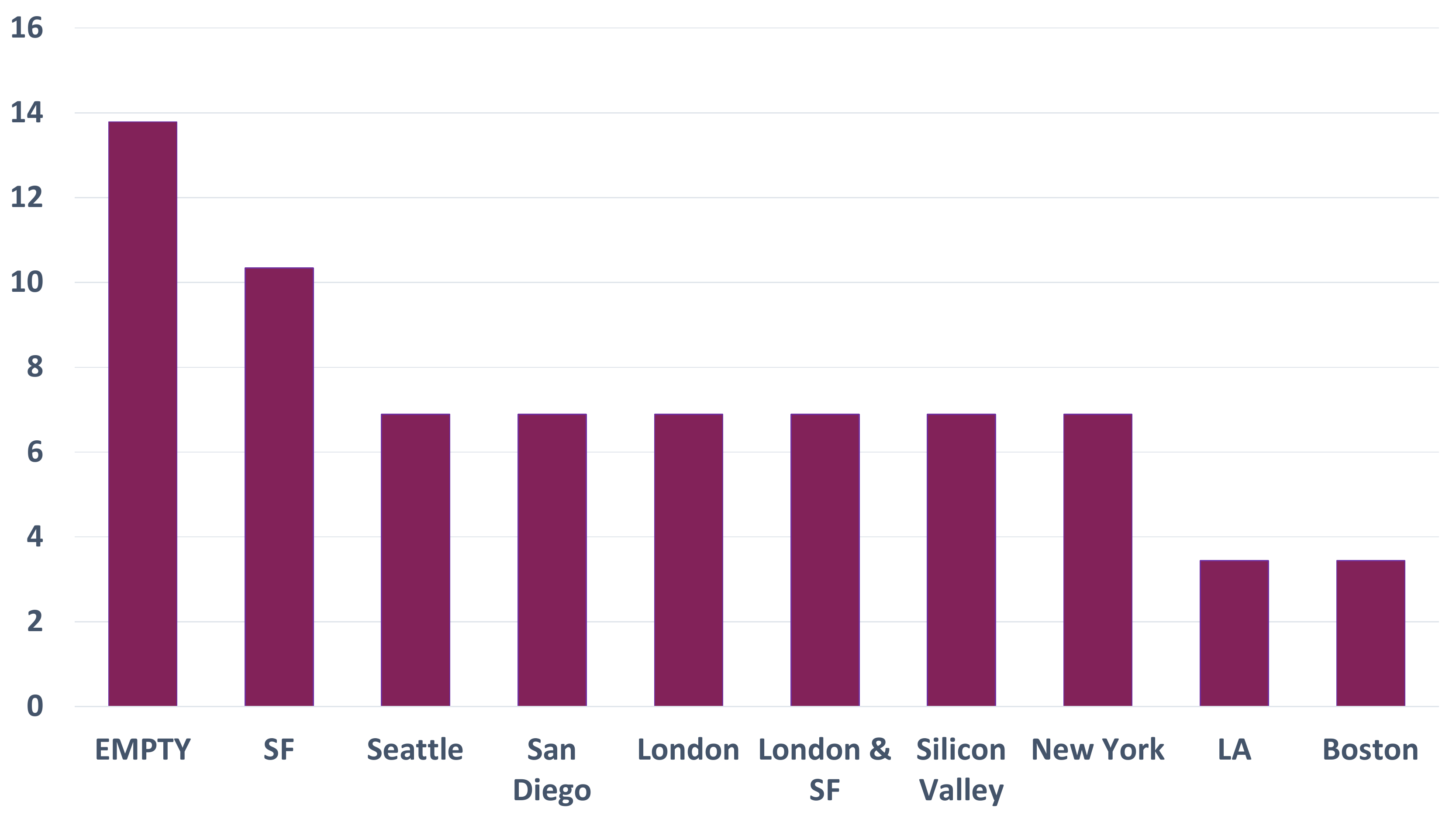}
        \caption{Experts}
    \end{subfigure}
\caption{Top-10 locations of users. Horizontal axis represents the location; vertical axis represents the percentage of users who are situated at a given location}
\label{fig:locations}
\end{figure}

Table~\ref{tab:dataStats} compares the statistics about the length of posts made by AIT and EAIT. On average, EAIT's tweets are longer compared to the tweets posted by AIT. We then looked in to the particulars of the users' geographical location and professional background which we obtained from their biography profiles. Figure~\ref{fig:locations} shows that the highest percentage of users in our dataset who tweet about AI are from London (6\% of the total set of users) followed by New York City (4\%) and Paris (3\%). The histogram shows only the top-10 locations of users as we also have a fair percentage of users tweeting from locations such as Sydney (1\%), Berlin (0.98\%), etc. Where as from EAIT, large percentage (14\%) of users in this set do not provide their geographical location. However, the top-2 locations of experts are San Francisco (10\%) followed by Seattle (7\%).

We conducted a unigram-based analysis of the profiles of the users to decipher their professional background. Table~\ref{tab:professions} shows that based on the frequencies of professions stated by users on Twitter, majority of the Twitter users contributing to AI-related tweets are pursuing careers in technology. 

\begin{table}[h]
\sffamily
\scriptsize
\centering
\setlength{\tabcolsep}{3pt}
\begin{tabular}{ | c | p{5.4cm} |} \hline
\textbf{User Category} & \textbf{Occupation} \\ \hline
AIT & entrepreneur, director, manager, enthusiast, founder, consultant, ceo, author, leader, engineer \\ \hline
EAIT & scientist, author, ceo, director, entreprenuer, cofounder, expert, researcher, technologist, professor \\ \hline
\end{tabular}
\caption{Top-10 occupations extracted from the user biographies}
\label{tab:professions}
\end{table}

\subsection{User Interests}
\label{sec:userana}
To examine the interests of the users in our dataset, we crawled the recent 100 posts made by these users. We extract the topics from these posts made by users using the Twitter LDA package~\cite{Zhao2011Traditional}. By utilizing these topics, we aim to measure the level of users' interest in technology which can quantify their perceptions about AI. 

\subsubsection{AIT}
As mentioned earlier, there are a total of 33K unique set of users who contributed towards our dataset. We crawl their recent tweets to extract the topics. We empirically decided to extract 5 topics and their corresponding vocabulary are shown as below: 

\begin{enumerate}
\item {\em \textbf{Topic-1} [Science \& Technology]: data, learning, intelligence, business, machine, digital, analytics, cloud, trends, science
\item \textbf{Topic-2} [Personal Status \& Opinions]: people, time, day, good, love, today, life, work, happy, story
\item \textbf{Topic-3} [General News]: latest, stories, daily, join, great, world, march, live, week, day, event, network, news
\item \textbf{Topic-4} [Non-English Tweets]: pour, les, des, para, los, dans, qui, avec, une, se, las
\item \textbf{Topic-5} [Daily Updates]: follow, daily, chapter, updates, translation, card }
\end{enumerate}

\begin{figure}[ht!]
\centering
\includegraphics[height=1.2in]{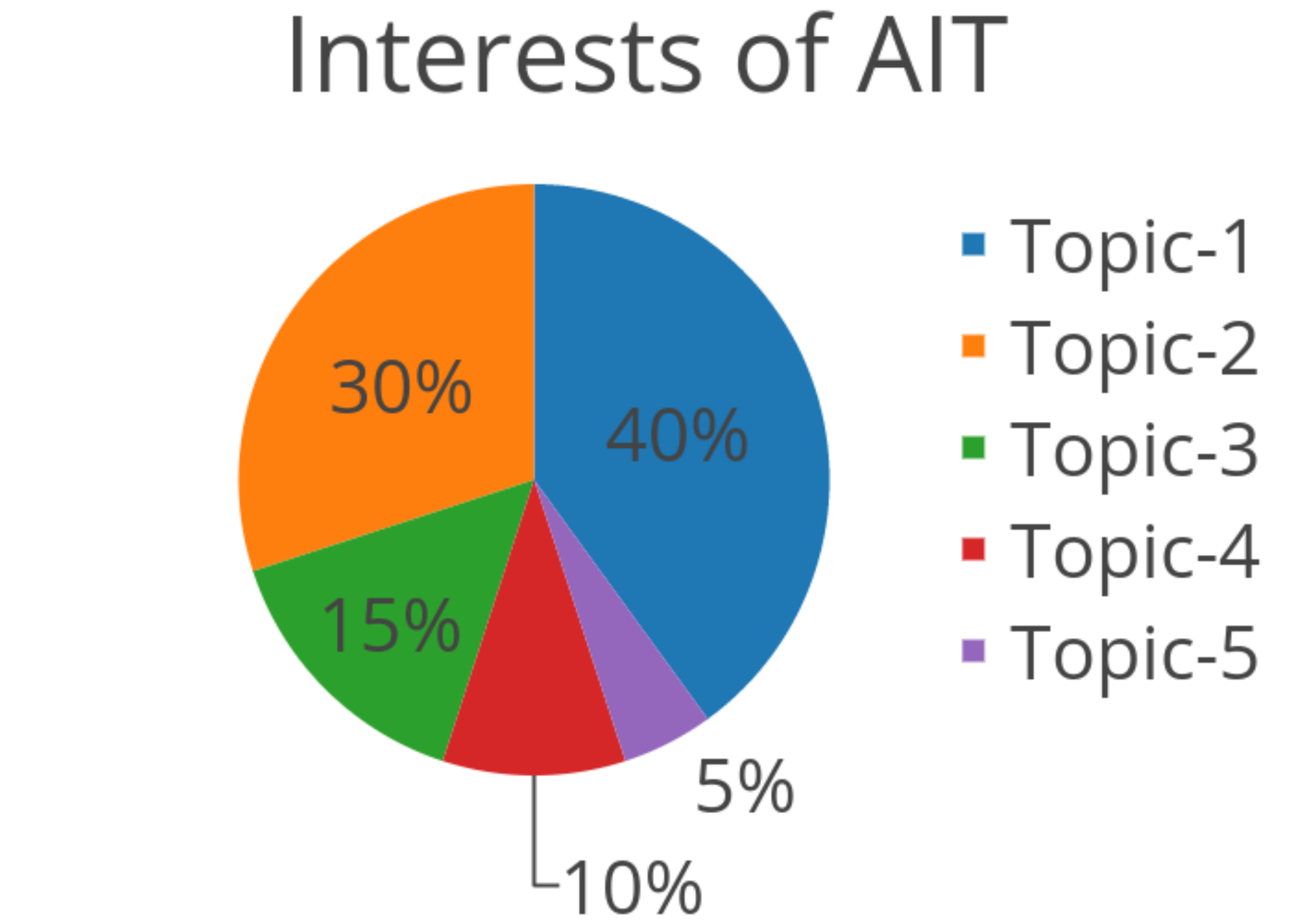}
\caption{Topic Distributions Extracted from Twitter. Topic-1: Science \& Technology; Topic-2: Personal Status \& Opinions; Topic-3: General News; Topic-4: Non-english Tweets; Topic-5: Daily Updates}
\label{fig:twittopne}
\end{figure}

By considering the topics, we obtain the percentage distribution of each individual's tweets to these 5 different topics. We then aggregate all the distributions of users across these topics and the percentage distributions are shown in Figure~\ref{fig:twittopne}. This figure shows that majority of the tweets posted by the users are about technology and science. 



\subsubsection{EAIT}
We conduct a similar investigation as above on the tweets posted by experts. The topics extracted and their corresponding vocabulary are shown here: 

\begin{enumerate}
\item {\em \textbf{Topic-1} [Rise of AI]: robots, intelligence, jobs, future, human, tesla, cars, driving, learning, machine 
\item \textbf{Topic-2} [AI subscriptions and research]: papers, news, subscribing, events, latest, learning, deep, code, work, image, model, people
\item \textbf{Topic-3} [AI News from industry]: zapchain, google, sharing, bitcoin, startup, slack, twitter, working, app, facebook, medium, chatbots
\item \textbf{Topic-4} [Marketing \& AI]: virtualreality, ibmwatson, team, marketing, virtual, work, channel, bit, time
\item \textbf{Topic-5} [Opinions about AI]: essays, tweeting, enjoy, visit, brain, neuroscience, conspiracy, cleaning, fiction, theory, fun, hype, post}
\end{enumerate}

\begin{figure}[ht!]
\centering
\includegraphics[height=1.2in]{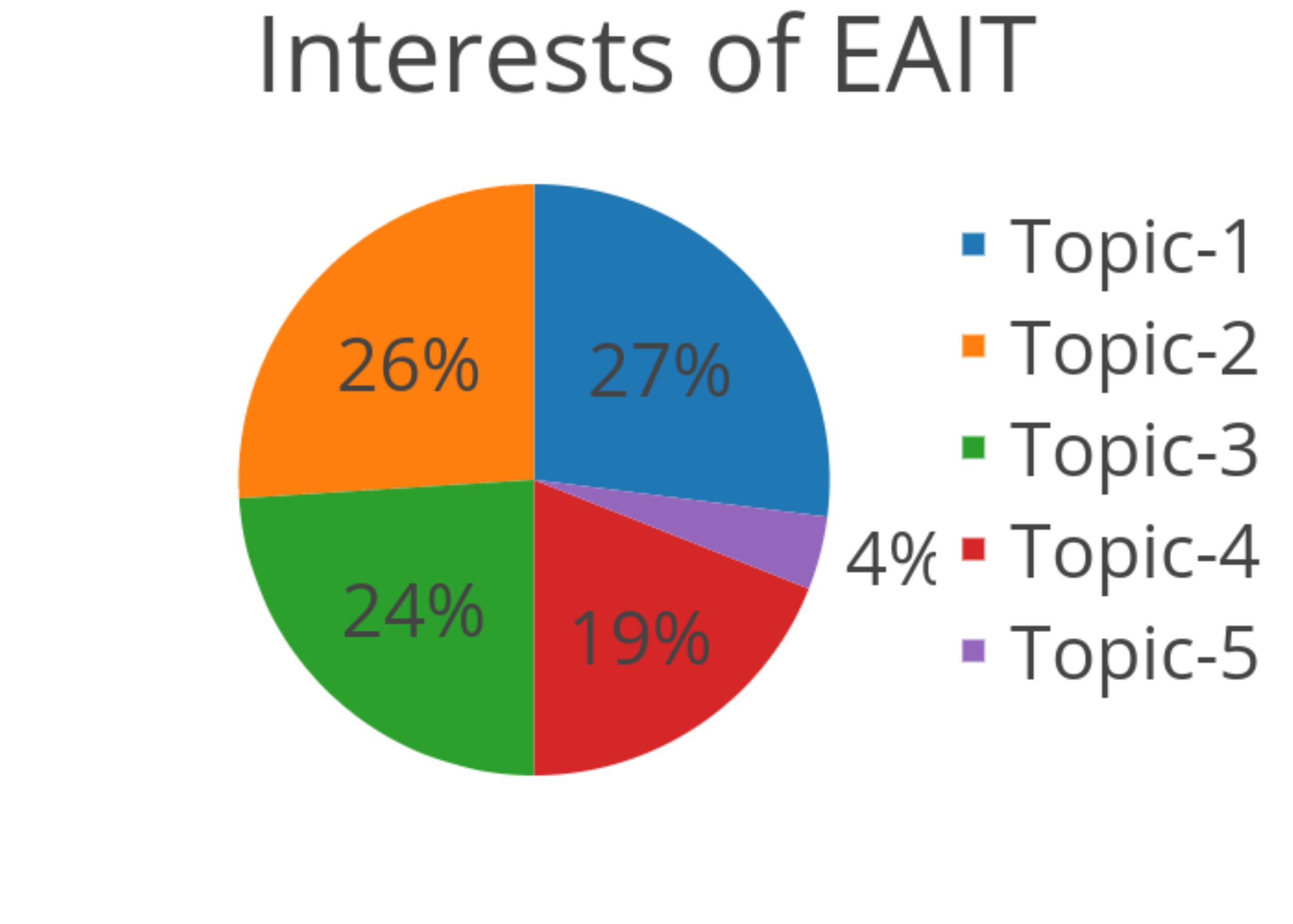}
\caption{Topic Distributions Extracted from Twitter. Topic-1: Rise and impact of AI; Topic-2: AI subscriptions and research; Topic-3: AI News from industry; Topic-4: Marketing \& AI; Topic-5: Opinions about AI}
\label{fig:twittope}
\end{figure}

These topics reveal that experts predominantly focus on posting about AI, its impacts, industry aspects of AI, their opinions and subscription recommendations which may help provide more information about AI. These topics are different from the topics focused by AIT.
The pie chart shown in Figure~\ref{fig:twittope} reveals that more than 50\% of the tweets posted by experts on Twitter are about the impacts of AI and research directions in this field. AIT share large percentage of personal opinions and statuses where as EAIT post the least percentage of tweets about their opinions. 

\section{RQ2: Twitter Engagement}

We seek to study the attributes which might disclose the holistic picture of the overall engagement rate of AI-related tweets. This rate provides us with an information on patterns of public interests and perceptions in AI. We measure the engagement by considering the `favorites', `likes', `replies' and `mentions' of a Twitter post. We first compute Twitter engagement statistics that are shown in Table~\ref{tab:usereng}.

\begin{table}[h]
\small
\centering
\setlength{\tabcolsep}{4pt}
\begin{tabular}{ |l|*{2}{c|}*{2}{c|}} \hline
 & \multicolumn{2}{c|}{Min (Max)} & \multicolumn{2}{c|}{Median (Mean)} \\ \hline 
 & AIT & EAIT & AIT & EAIT \\ \hline
Retweets & 0 (3364) & 0 (31297) & 2.0 (19.38) & 0.0 (39.61)\\ \hline
Favorites & 0 (69) & 0 (109569) & 0.0 (0.12) & 1.0 (97.79)\\ \hline
Mentions & 0 (10) & 0 (12) & 1.0 (1.05) & 1.0 (0.97)\\ \hline
\end{tabular}
\caption{Min (Max) and Median (Mean) values of Retweets, Favorites, Mentions}
\label{tab:usereng}
\end{table}

Tweets made by AIT are more likely to be retweeted than favorited by the users on this platform. 63.3\% of their tweets are retweeted atleast once. This is significantly higher than the general dataset which is 11.99\% as shown by the existing literature~\cite{Suh2010socom}. Where as, the tweets posted by EAIT received more number of favorites than retweets as only 35\% of the tweets are retweeted atleast once. 

Tweets from both the categories of users have higher probabilities of containing atleast one user handle. This may suggest that users are more likely to interact or engage in discussions with each other about AI on Twitter. On the other hand, 68.5\% of the tweets posted by AIT in our dataset have atleast one url shared as part of the tweet where as 48\% of the tweets made by EAIT have urls. Sharing large percentage of urls by AIT compared to EAIT could be one of the reasons for receiving more retweets than favorites. Literature~\cite{Java2007,Kwak2010,yang2010predicting} considers retweeting as one of the features to measure information diffusion. Based on these results, tweets posted by AIT diffuse faster (higher retweet rate) than the tweets posted by EAIT.

\section{RQ3: Optimistic or Pessimistic}
Optimism is defined as being hopeful and confident about the future whose synonym is `positive' and pessimism is defined as tending to see or believing that the worst will happen whose synonym is `negative'. In this work we measure these two attributes -- positive, negative emotions alongside assessing cognitive mechanisms by employing the popular psycho-linguistic tool LIWC~\cite{TausczikY2010}. Tausczik et. al in their work introducing LIWC mention that the way people express emotion and the degree to which they express it can tell us how people are experiencing the world. Existing literature~\cite{Danescu2011words,Tsur2012wsdm,Tumasjan2010Twitter} states that LIWC is powerful in accurately identifying emotion in the usage of language. This is the motivation for us to use LIWC to measure the emotionality of tweets shared by AIT and EAIT.


\subsection{AIT}
\begin{figure*}[!ht]
        \centering
    \begin{subfigure}[t]{0.3\textwidth}
        \centering
        \includegraphics[height=1.2in]{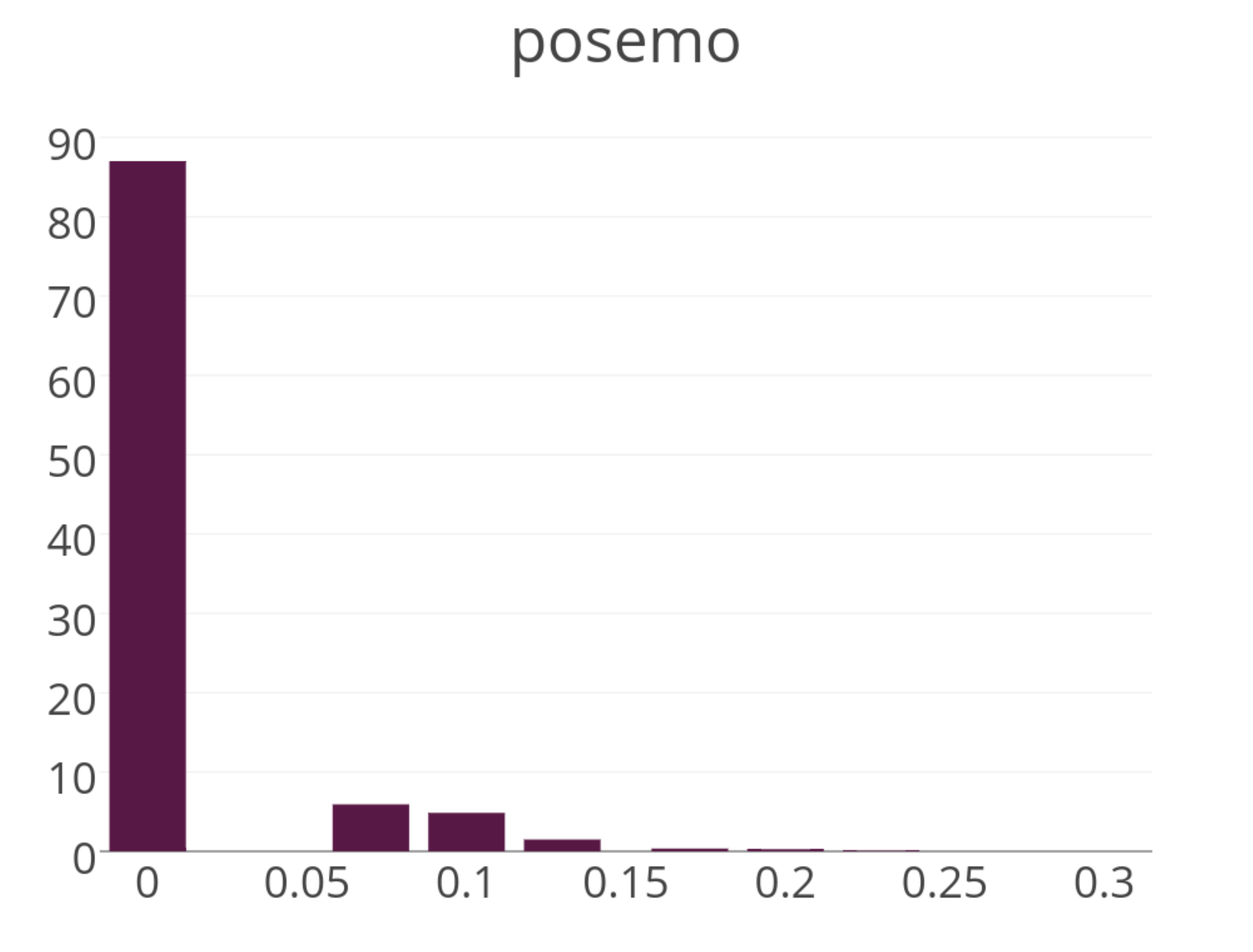}
	\caption{Positive Emotions}
    \end{subfigure}%
    ~ 
    \begin{subfigure}[t]{0.3\textwidth}
        \centering
        \includegraphics[height=1.2in]{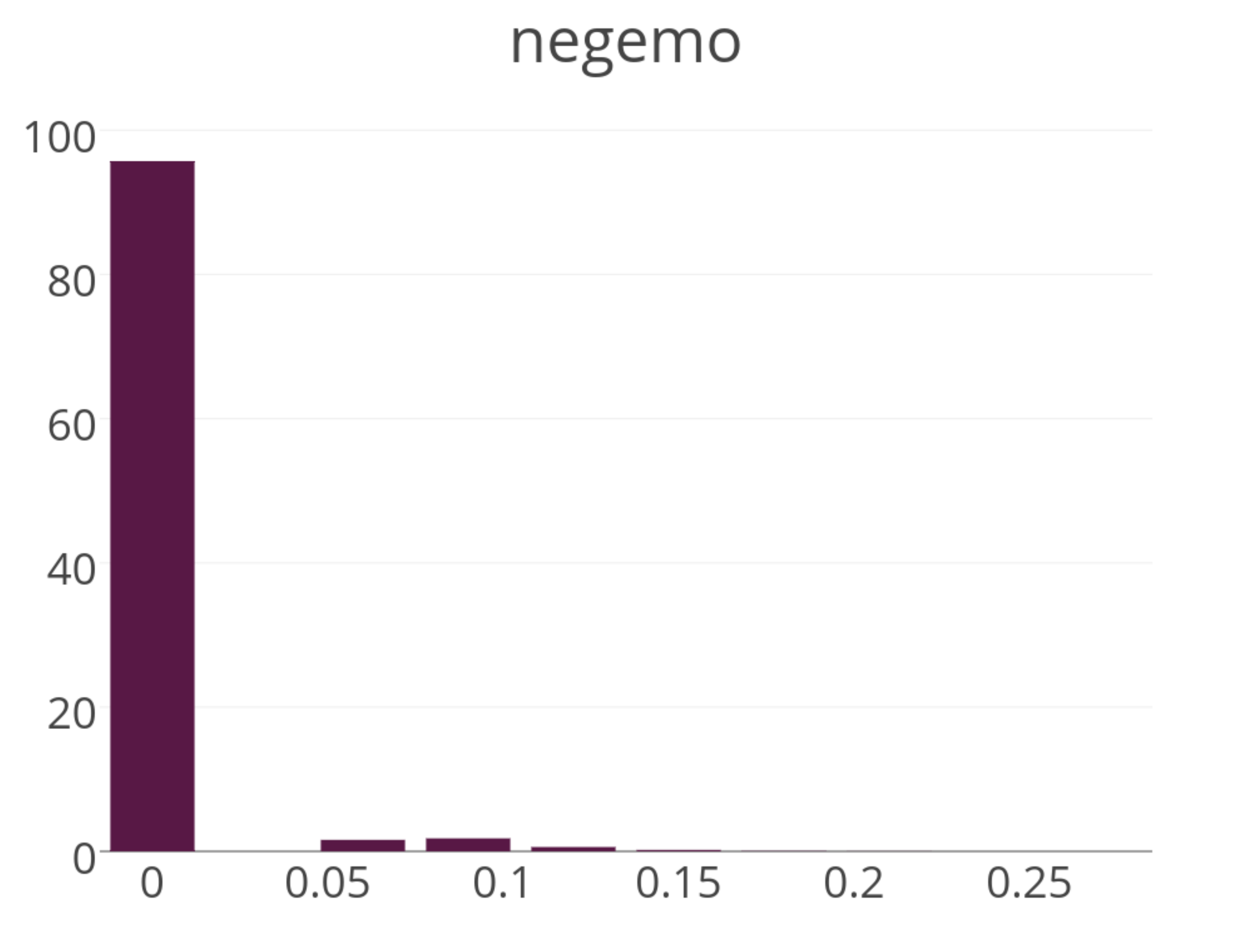}
	\caption{Negative Emotions}
    \end{subfigure}
    ~ 
    \begin{subfigure}[t]{0.3\textwidth}
        \centering
        \includegraphics[height=1.2in]{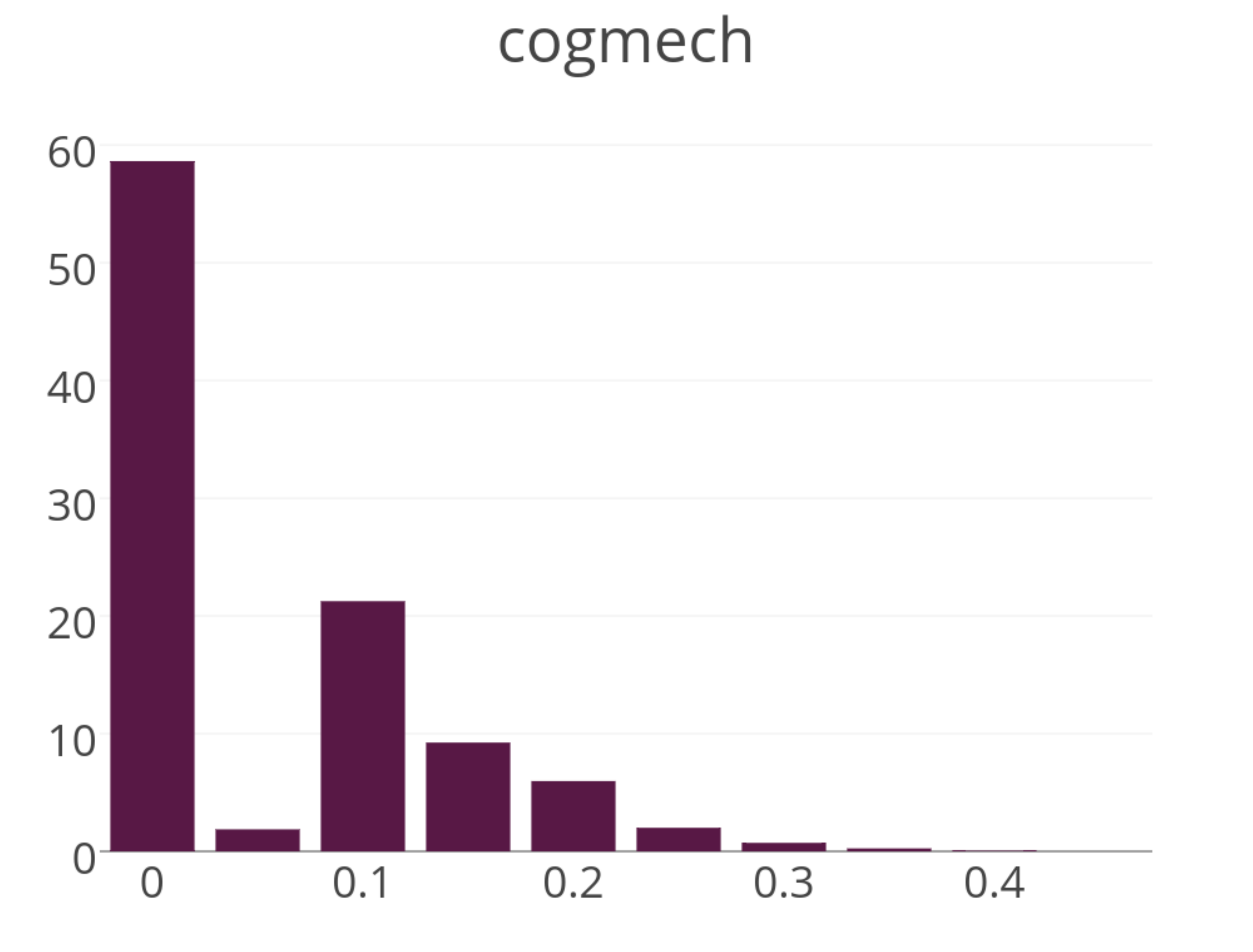}
	\caption{Insights}
    \end{subfigure}
    \caption{Emotions from the tweets posted by AIT}
    \label{fig:liwcnexp}

\end{figure*}

Figure~\ref{fig:liwcnexp} reveals that individuals are more postive (65\% greater than negative) and optimistic towards AI and its related topics. This concurs with the recent analysis~\cite{EthanAAAI2016} and survey~\cite{LeslieHarvard2016} on New York Times articles and interviews with individuals respectively. This is a useful finding because Twitter is known for users making posts that are emotionally negative~\cite{LydiaICWSM2016}. In other words, despite the general negative emotional content on Twitter, thes subset of tweets focusing on artificial intelligence are more positive than being negative.


\subsection{EAIT}
\begin{figure*}[!ht]
\centering
    \begin{subfigure}[t]{0.3\textwidth}
        \centering
        \includegraphics[height=1.2in]{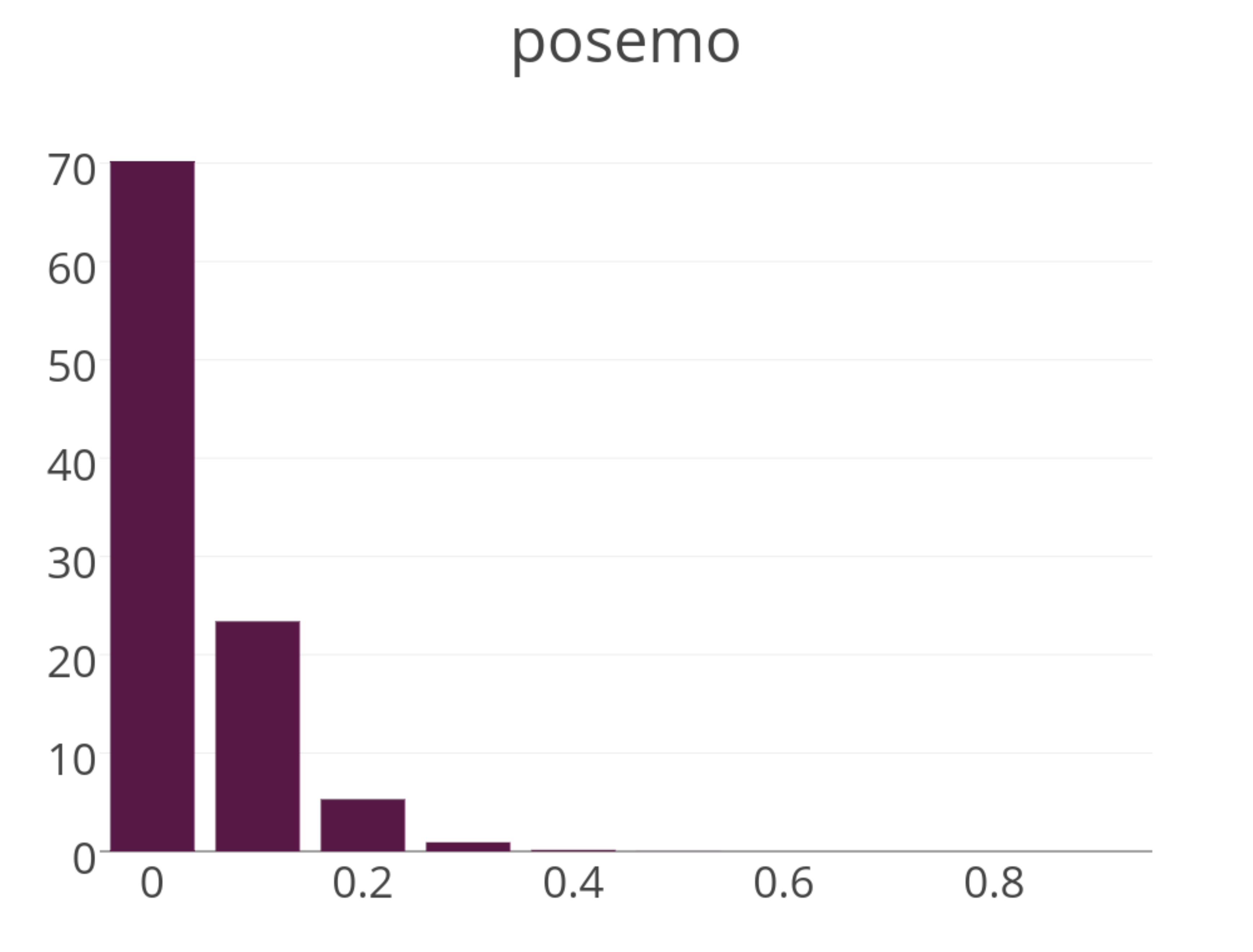}
	\caption{Positive Emotions}
    \end{subfigure}%
    ~ 
    \begin{subfigure}[t]{0.3\textwidth}
        \centering
        \includegraphics[height=1.2in]{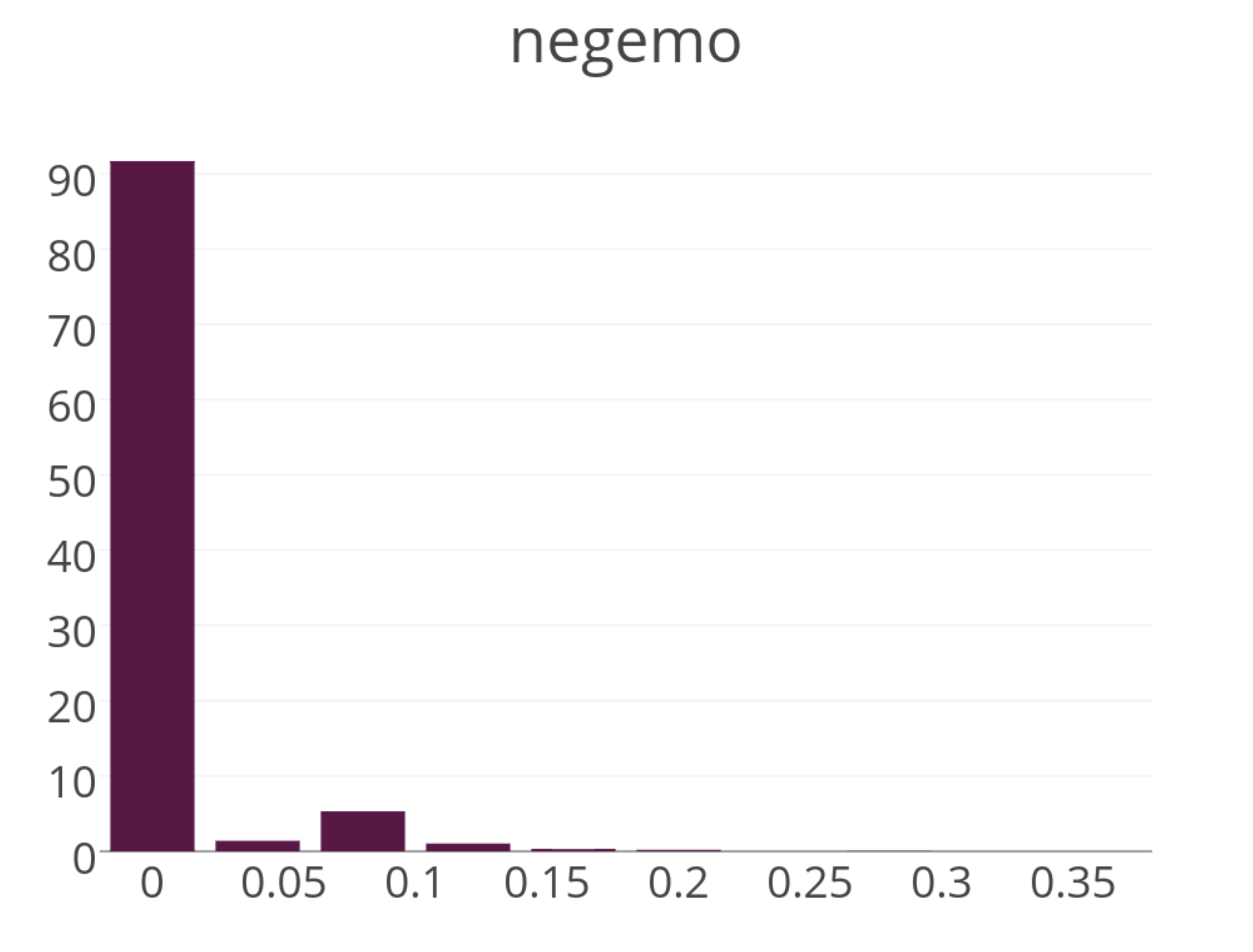}
	\caption{Negative Emotions}
    \end{subfigure}
    ~ 
    \begin{subfigure}[t]{0.3\textwidth}
        \centering
        \includegraphics[height=1.2in]{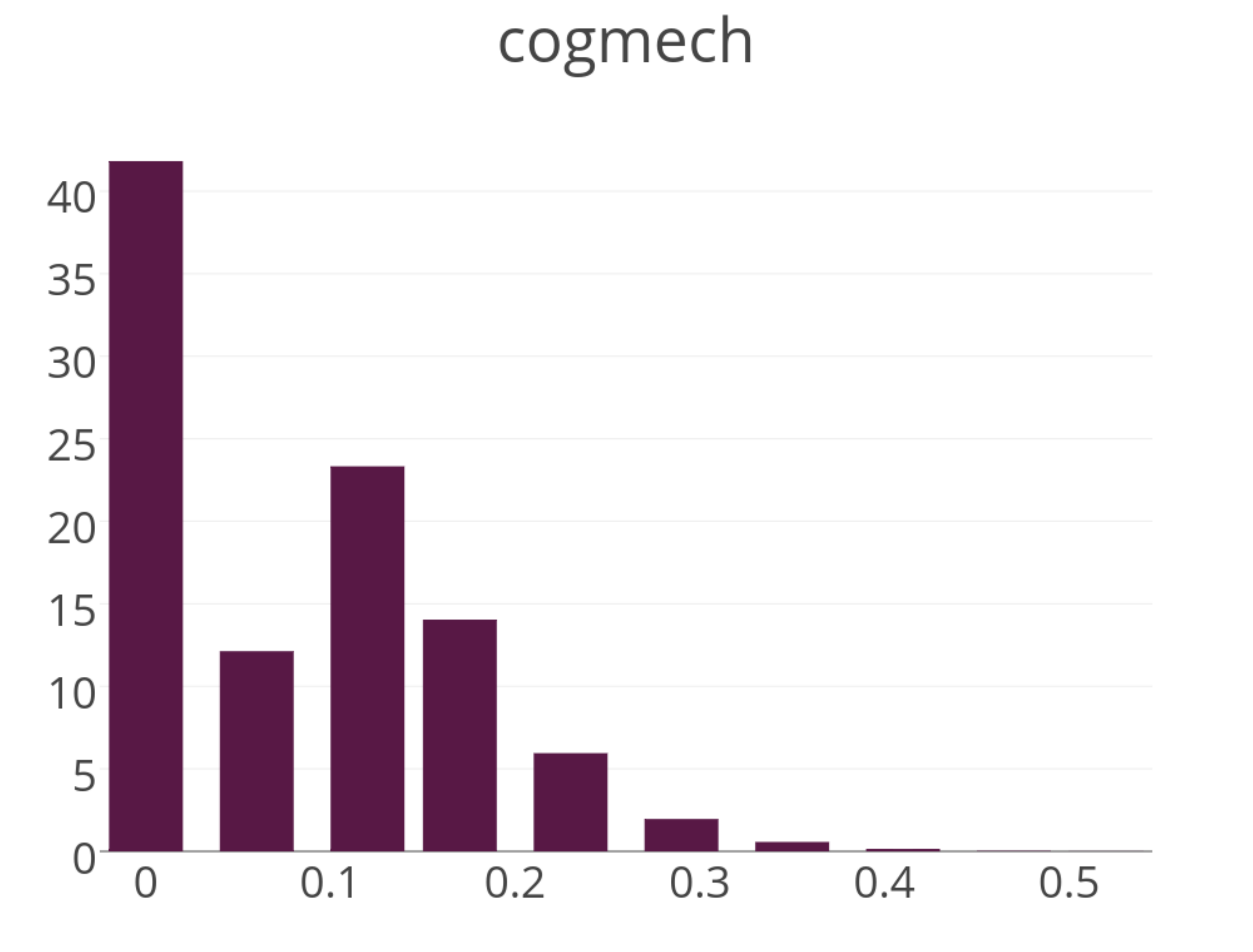}
	\caption{Insights}
    \end{subfigure}
    \caption{Emotions from the tweets posted by EAIT}
    \label{fig:liwcexp}
\end{figure*}

We conduct the emotion analysis on tweets posted by experts and it reveals similar findings as earlier (shown in Figure~\ref{fig:liwcexp}) but with relatively higher cognitive mechanisms than AIT. Cognitive mechanisms or complexity can be considered as a rich way of reasoning~\cite{TausczikY2010}. The horizontal axis represents the gravity of a given emotion and the vertical axis represents the number of posts with a gravity value shown on horizontal axis. The distribution in each plot are normalized and the sum of all the values in different buckets of gravity will sum upto 100\%.

\vspace{2in}
\noindent
Compared to AIT, tweets made by EAIT have more negativity overall. However, positive emotion is thrice as dominating as the negative emotion. When we compare the positive and negative emotions of the two categories of users, the results reveal that expert users are 38\% more negative than the general AI-Tweeters.

\section{RQ4: Topics heavily discussed by users on Twitter}
In Section~\ref{sec:userana}, we have presented the analysis on the interests of users by crawling their timelines and extracting topics from these timelines. In order to better understand the public perceptions about AI, we extract topics from the tweets that are exclusively about AI. 

To perform this, we first consider all the tweets posted by AIT using the hashtag approach. In Table~\ref{tab:topicvocab}, we present the topics extracted by considering the aggregated set of tweets. These topics display that the focus of the AI-tweets are on stories and statistics about AI followed by discussions about deep learning. These topics reveal the high-level interests of individuals about the emerging trends, impacts and career opportunities related to AI. 

These topics also display that individuals have continued interests in the similar topics over years due to partial alignment of these topics with the findings shown by Fast et. al~\cite{EthanAAAI2016} who consider NYT articles published before 2017. We perform this analysis on the AI-tagged tweets from Twitter as we assume that this set may comprise both experts and general Twitter users. 

\begin{table*}[!ht]
\centering
\small
\begin{tabular}{l | l |  p{7.4cm} | l }  
\textbf{ID} & \textbf{Topic} & \textbf{Top Tags} & \textbf{\% of tweets} \\ \hline
1 & Stories \& Statistics & future, stories, problems, statistics, analytics & 13.8\% \\ \hline
2 & Deep learning & deeplearning, deep, neuralnetworks, tensorflow, human & 12.2\% \\ \hline
3 & Emerging trends & trends, 2017, iot, machinelearning, vr, ar & 11.7\% \\ \hline
4 & Applications & cars, robotics, healthcare, startups, banking & 11.43\% \\ \hline
5 & Lessons \& learnings & guide, cloud, healthcare, lessons, rstats & 10.4\% \\ \hline
6 & Impact \& predictions & chatbots, world, impact, influence, predicts & 8.87\% \\ \hline
7 & Applications \& impact of AI & chatbots, cancer, robots, jobs, automation & 8.5\% \\ \hline
8 & Jobs \& recruitment & experts, marketing, hiring, consulting, experience & 8.3\% \\ \hline
9 & Business \& revenue & innovation, growth, companies, revenue, startup & 7.6\% \\ \hline
10 & Big data \& data science & analytics, bigdata, datascience, innovation, smart & 7.2\% \\ \hline
\end{tabular}
\caption{10 topics and the corresponding vocabulary extracted along with the percentage distribution of tweets across these topics.}
\label{tab:topicvocab}
\end{table*}


\section{RQ5: Co-occurring concepts discussed}
The questions we investigated until now provides valuable insights into whether and how individuals perceive the issues about AI advancements. However, we note that conceptual relationships could significantly quantify and measure the perceptions of individuals. Towards addressing this challenge, we employ the popular word2vec analysis to detect relationships between words that are frequently co-occurring. Word2Vec~\cite{TomasWord13} is a popular two-layer neural network that is used to process text. It considers a text corpus as an input and generates feature vectors for words present in that corpus. Word2vec represents words in a higher-dimensional feature space and makes accurate predictions about the meaning of a word based on its past occurrences. These vectors can then be utilized to detect relationships between words which are highly accurate given enough data to learn these vectors. 

\begin{figure}[!ht]
\centering
    \begin{subfigure}{0.2\textwidth}
        \centering
	\includegraphics[scale=0.24]{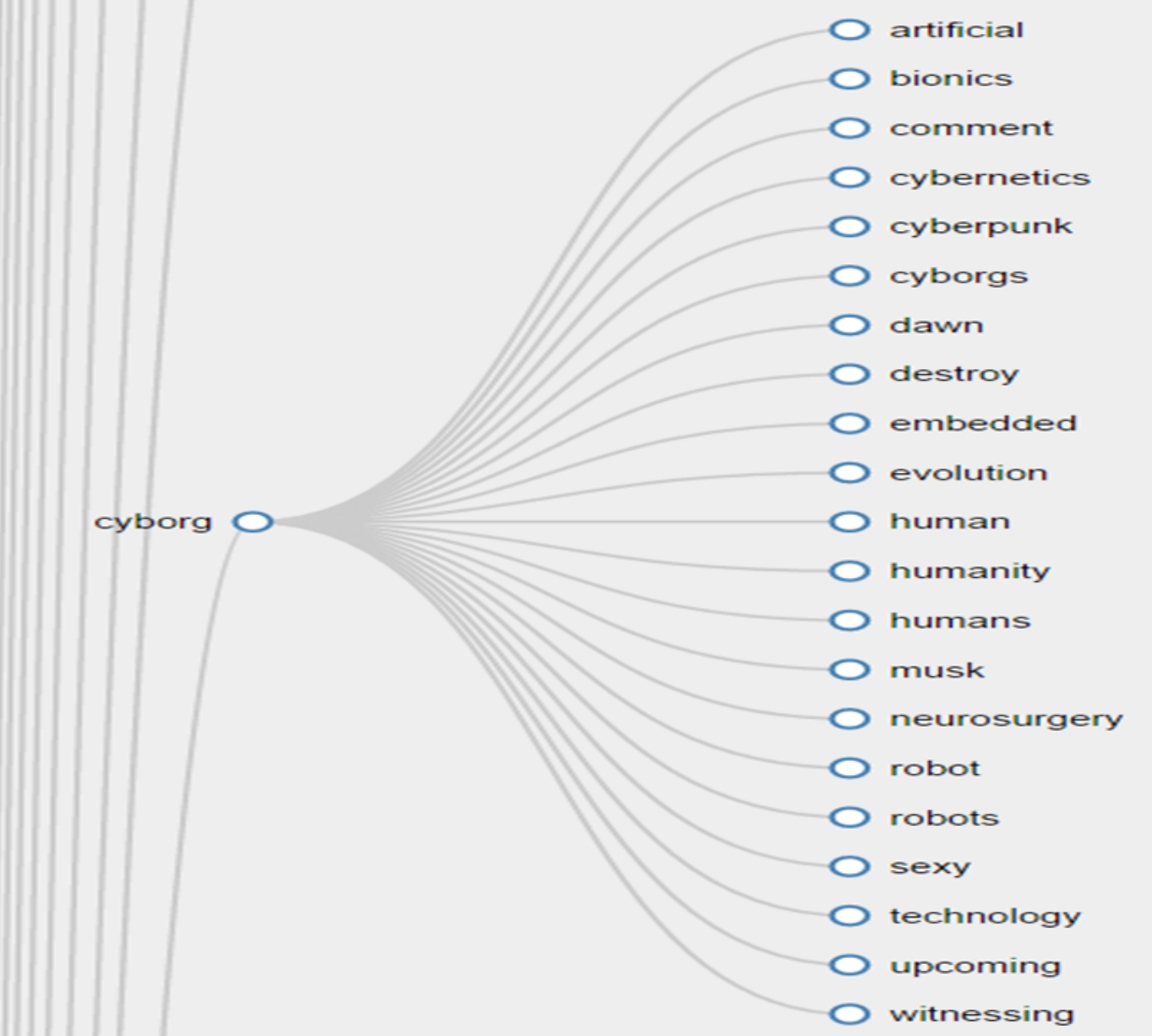}
	\caption{Cyborg--AIT}
    \end{subfigure}
    ~
    \begin{subfigure}{0.2\textwidth}
	\centering
	\includegraphics[scale=0.24]{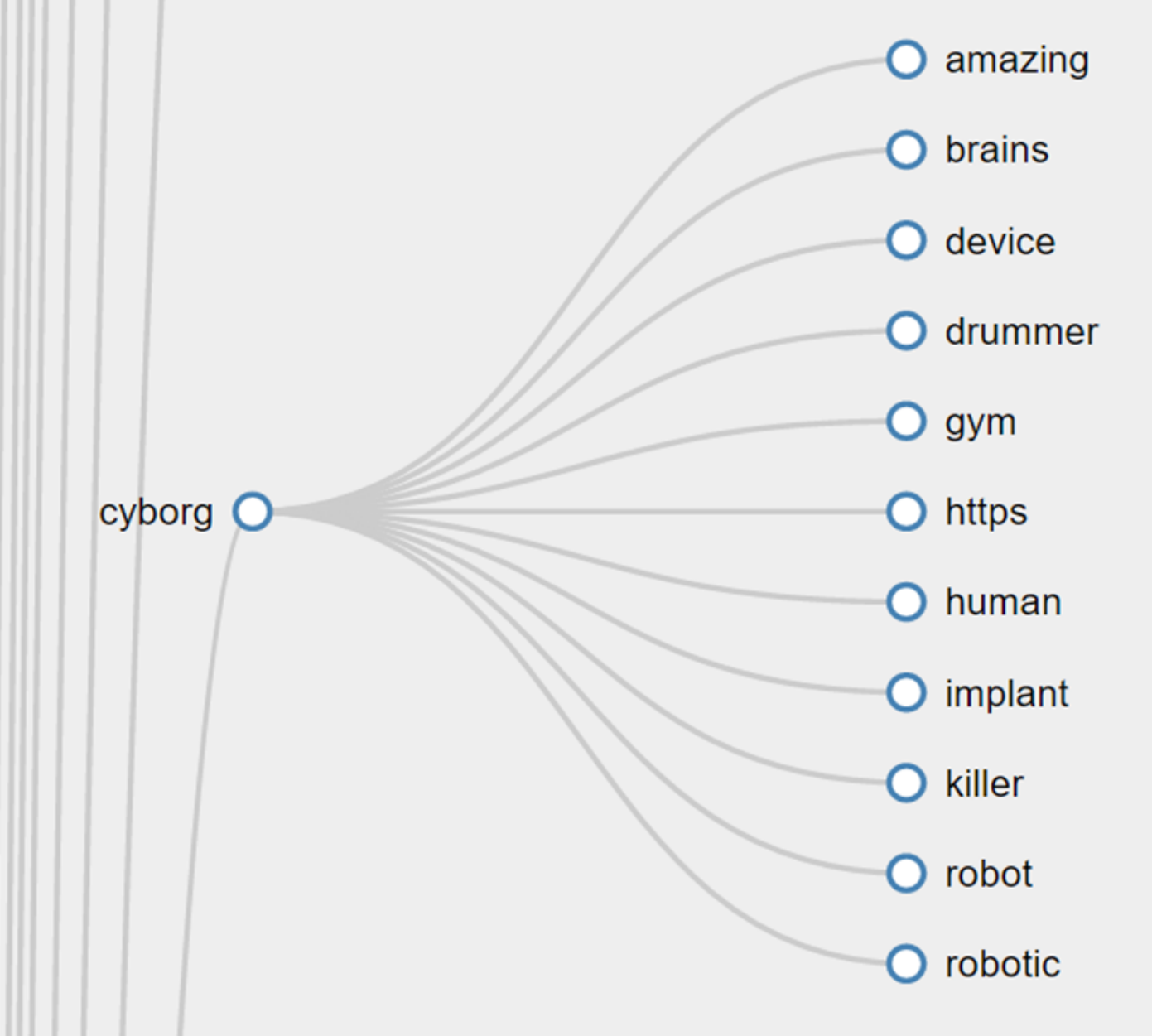}
	\caption{Cyborg--EAIT}
    \end{subfigure}
    ~
    \begin{subfigure}{0.2\textwidth}
        \centering
	\includegraphics[scale=0.22, width=1.3in]{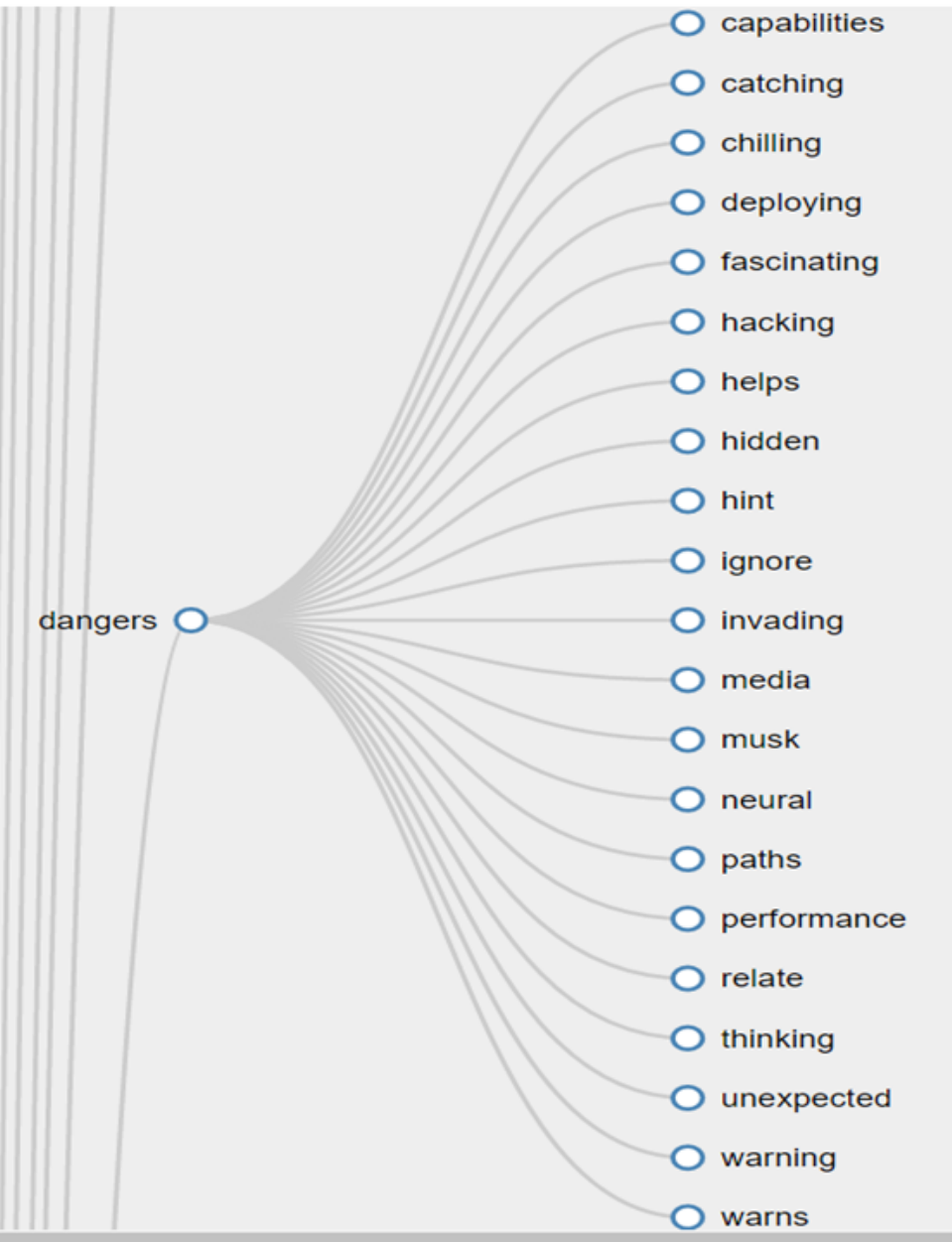}
	\caption{Dangers--AIT}
    \end{subfigure}
    ~
    \begin{subfigure}{0.2\textwidth}
	\centering
	\includegraphics[scale=0.22,width=1.3in]{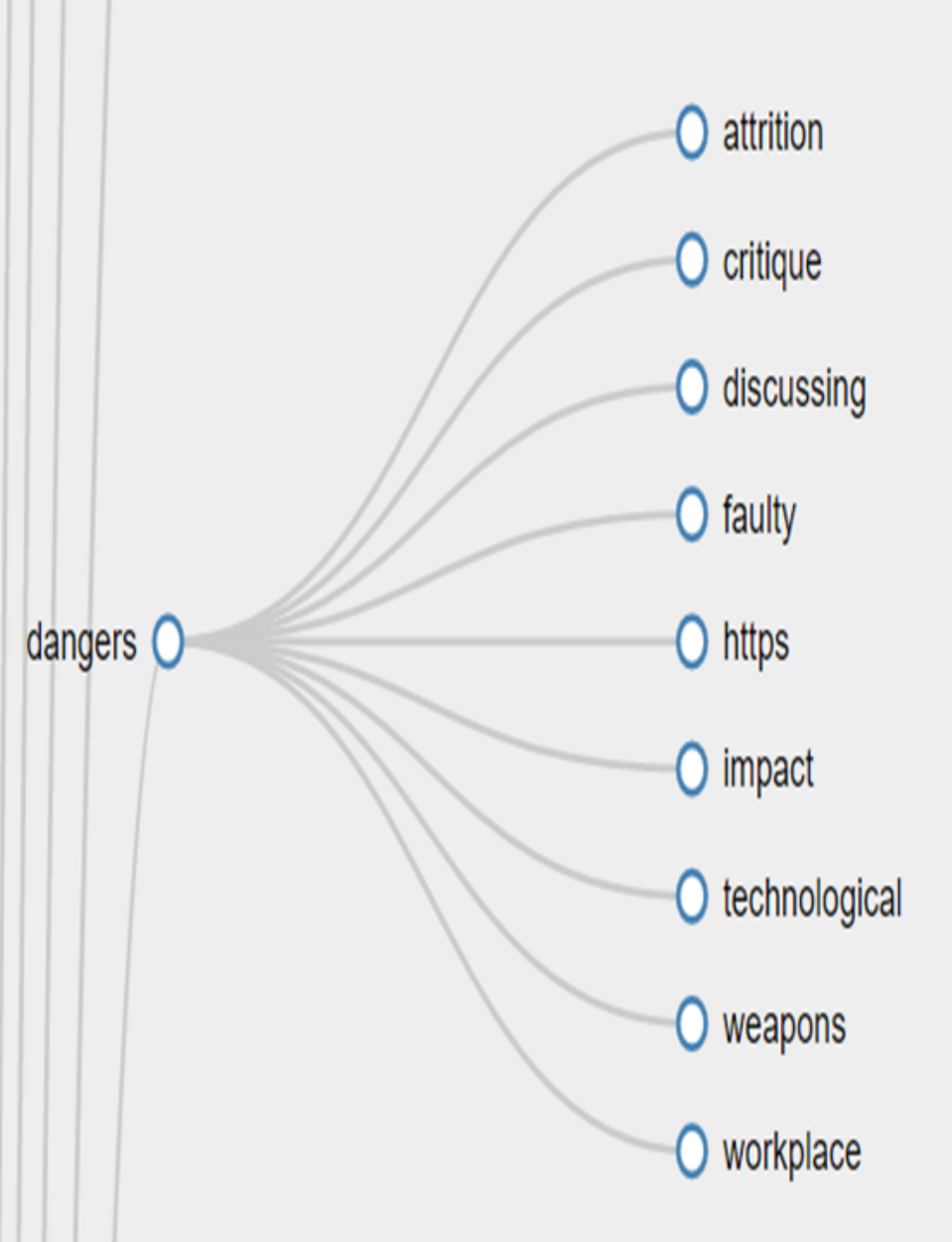}
	\caption{Dangers--EAIT}
    \end{subfigure}
    \caption{Co-occurring patterns. The word on the left belongs to the AI vocabulary which is mapped to multiple words that are used by the users in their tweets about AI.}
    \label{fig:cooccur}
\end{figure}

To detect the relationships, we train the Word2Vec model on the \emph{Text8} corpus (\url{http://mattmahoney.net/dc/}) which is created using the articles from Wikipedia. As a processing step, we first remove stop words from the tweets and consider each tweet independently. We utilized the pre-existing lists from academia\footnote{https://www.cs.utexas.edu/users/novak/aivocab.html} and industry\footnote{http://www.techrepublic.com/article/mini-glossary-ai-terms-you-should-know/} to manually compile the AI vocabulary of 61 words. Figure~\ref{fig:cooccur} provides two pairs of co-occurring pattern comparisons. These pairs of co-occurring patterns tell us that AIT are in general fantasizing about the future where as EAIT are grounded and realistic. The entire list of co-occurrences for the 61 words in AI vocabulary can be found here: AIT (\url{https://goo.gl/5WPA9L}) and EAIT (\url{https://goo.gl/8WxBaP}).


\section{Conclusions}
Social media platforms are one of the primary channels of communication in the lives of individuals. These platforms are reshaping our ideas and the way we share those ideas. Given the increasing interest in AI from different communities, multiple debates are commencing to evaluate the benefits and drawbacks of AI to humans and society as a whole. This paper presents the findings from our investigation on public perceptions about AI using the AI-related posts shared on Twitter. Alongside, we performed a comparative analysis between how the posts made by AIT and EAIT are engaged. Some of the key findings from our analysis are: 

\begin{enumerate}

\item Tweets about AI are overall more positive compared to the general tweets. 
\item Tweets posted by experts are more negative than the general AI tweeters. 
\item Tweets posted by experts have lower diffusion than the tweets posted by general AI tweeters. 
\item General AI-tweeters are geographically distributed with London and New York City as the top locations. 
\end{enumerate}

The co-occurring pattern mapping tells us that users belonging to EAIT are more grounded and realistic in their perceptions about AI. Additionally, analysis on user interests revealed that tweeters who are posting about AI are interested in technology. The most discussed topics on Twitter by the general AI tweeters are about the \emph{stories and statistics} alongside the impact of deep learning, career and recruitment aspects. Our LIWC analysis revealed that discussions are cognitively loaded with a larger variance of gravities among Twitter users.

We hope that our findings will benefit different organizations and communities who are debating about the benefits and threats of AI to our society. Some of the future directions include a longitudinal study across several years as well as multiple mediums of communication. 

\subsection{Acknowledgements}
This research is supported in part by a Google research award, the ONR grants N00014-16-1-2892, N00014-13-1-0176, N00014-13-1-0519, N00014-15-1-2027 and the NASA grant NNX17AD06G. We thank Miles Brundage for his feedback on parts of this analysis.

\bibliographystyle{aaai}
\bibliography{references}

\begin{thebibliography}{}

\bibitem[\protect\citeauthoryear{Bakshy \bgroup et al\mbox.\egroup
  }{2011}]{Bakshy2011EIQ}
Bakshy, E.; Hofman, J.~M.; Mason, W.~A.; and Watts, D.~J.
\newblock 2011.
\newblock Everyone's an influencer: Quantifying influence on twitter.
\newblock In {\em Proceedings of the Fourth ACM International Conference on Web
  Search and Data Mining}, WSDM '11.

\bibitem[\protect\citeauthoryear{Danescu-Niculescu-Mizil, Gamon, and
  Dumais}{2011}]{Danescu2011words}
Danescu-Niculescu-Mizil, C.; Gamon, M.; and Dumais, S.
\newblock 2011.
\newblock Mark my words!: Linguistic style accommodation in social media.
\newblock In {\em Proceedings of the 20th International Conference on World
  Wide Web}.

\bibitem[\protect\citeauthoryear{Fast and Horvitz}{2016}]{EthanAAAI2016}
Fast, E., and Horvitz, E.
\newblock 2016.
\newblock Long-term trends in the public perception of artificial intelligence.
\newblock In {\em AAAI}.

\bibitem[\protect\citeauthoryear{Gaines-Ross}{2016}]{LeslieHarvard2016}
Gaines-Ross, L.
\newblock 2016.
\newblock What do people -- not techies, not companies -- think about
  artificial intelligence?
\newblock In {\em {H}arvard {B}usiness {R}eview}.

\bibitem[\protect\citeauthoryear{Java \bgroup et al\mbox.\egroup
  }{2007}]{Java2007}
Java, A.; Song, X.; Finin, T.; and Tseng, B.
\newblock 2007.
\newblock Why we twitter: Understanding microblogging usage and communities.
\newblock In {\em Proceedings of the 9th WebKDD and 1st SNA-KDD 2007 Workshop
  on Web Mining and Social Network Analysis}, WebKDD/SNA-KDD '07.

\bibitem[\protect\citeauthoryear{Kwak \bgroup et al\mbox.\egroup
  }{2010}]{Kwak2010}
Kwak, H.; Lee, C.; Park, H.; and Moon, S.
\newblock 2010.
\newblock What is twitter, a social network or a news media?
\newblock In {\em Proceedings of the 19th International Conference on World
  Wide Web}, WWW '10.

\bibitem[\protect\citeauthoryear{Manikonda, Meduri, and
  Kambhampati}{2016}]{LydiaICWSM2016}
Manikonda, L.; Meduri, V.~V.; and Kambhampati, S.
\newblock 2016.
\newblock Tweeting the mind and instagramming the heart: Exploring
  differentiated content sharing on social media.
\newblock In {\em International AAAI Conference on Web and Social Media}.

\bibitem[\protect\citeauthoryear{Mikolov \bgroup et al\mbox.\egroup
  }{2013}]{TomasWord13}
Mikolov, T.; Chen, K.; Corrado, G.; and Dean, J.
\newblock 2013.
\newblock Efficient estimation of word representations in vector space.
\newblock {\em CoRR} abs/1301.3781.

\bibitem[\protect\citeauthoryear{Naaman, Boase, and Lai}{2010}]{Naaman2010}
Naaman, M.; Boase, J.; and Lai, C.-H.
\newblock 2010.
\newblock Is it really about me?: Message content in social awareness streams.
\newblock In {\em Proceedings of the 2010 ACM Conference on Computer Supported
  Cooperative Work}, CSCW '10.

\bibitem[\protect\citeauthoryear{Suh \bgroup et al\mbox.\egroup
  }{2010}]{Suh2010socom}
Suh, B.; Hong, L.; Pirolli, P.; and Chi, E.~H.
\newblock 2010.
\newblock Want to be retweeted? large scale analytics on factors impacting
  retweet in twitter network.
\newblock In {\em Proceedings of the 2010 IEEE Second International Conference
  on Social Computing}, SOCIALCOM '10.

\bibitem[\protect\citeauthoryear{Tausczik and Pennebaker}{2010}]{TausczikY2010}
Tausczik, Y.~R., and Pennebaker, J.~W.
\newblock 2010.
\newblock The psychological meaning of words: Liwc and computerized text
  analysis methods.
\newblock {\em Journal of Language and Social Psychology} 29(1):24--54.

\bibitem[\protect\citeauthoryear{Tsur and Rappoport}{2012}]{Tsur2012wsdm}
Tsur, O., and Rappoport, A.
\newblock 2012.
\newblock What's in a hashtag?: Content based prediction of the spread of ideas
  in microblogging communities.
\newblock In {\em Proceedings of the Fifth ACM International Conference on Web
  Search and Data Mining}.

\bibitem[\protect\citeauthoryear{Tumasjan \bgroup et al\mbox.\egroup
  }{2010}]{Tumasjan2010Twitter}
Tumasjan, A.; Sprenger, T.; Sandner, P.; and Welpe, I.
\newblock 2010.
\newblock Predicting elections with twitter: What 140 characters reveal about
  political sentiment.

\bibitem[\protect\citeauthoryear{Yang and Counts}{2010}]{yang2010predicting}
Yang, J., and Counts, S.
\newblock 2010.
\newblock Predicting the speed, scale, and range of information diffusion in
  twitter.
\newblock {\em ICWSM} 10:355--358.

\bibitem[\protect\citeauthoryear{Zhao \bgroup et al\mbox.\egroup
  }{2011}]{Zhao2011Traditional}
Zhao, W.~X.; Jiang, J.; Weng, J.; He, J.; Lim, E.-P.; Yan, H.; and Li, X.
\newblock 2011.
\newblock Comparing twitter and traditional media using topic models.
\newblock In {\em Proceedings of the 33rd European Conference on Advances in
  Information Retrieval}, ECIR'11.

\end{thebibliography}

\clearpage
\appendix

\section{Appendix -- Analysis on Reddit}
Reddit is one of the popular online social media platforms that is considered as a social news and a discussion platform. Users on this platform can share posts that may contain either text or media. Other users on this platform can then vote or comment on these posts. There are different sub-communities organized by the areas of interest also called as Subreddits. Posts on any subreddit are of two types -- self post or a link post. Self post is defined as a post that contains text and the content of this information is saved on the Reddit's server. Link post submission redirects the user to an external website. Similar to the Twitter analysis, we consider the subreddits that focus on AI to capture user perceptions about AI. To start the analysis, we match the hashtags used in Twitter analysis to the names of subreddits to conduct a comparative study between Reddit and Twitter. Also, please note that all the users considered for Reddit analysis are categorized as common Reddit users with no distinction between experts and non-experts.

\subsection{Data Collection}
To crawl the data from Reddit, we employed the Python Reddit API Wrapper~\footnote{https://praw.readthedocs.io/en/latest/} (PRAW). We utilized the subreddits that focus on the same topics as the hashtags used to crawl the data from Twitter -- \emph{r/artificial}, \emph{r/machinelearning} and \emph{r/datascience}. Across the 3 subreddits, a total of 2,550 unique thread posts are crawled from 5 different categories of posts -- \emph{hot}, \emph{new}, \emph{rising}, \emph{controversial} and \emph{top}. The total number of posts aggregated across all the threads consists of 21,420 posts.

For each post crawled using PRAW, the metadata includes the title of the post, content of the post, score of the post, up votes, down votes, posting date, username of the author. In our dataset, we have 5,382  unique number of users. For understanding the users and their background, we separately crawl the recent 100 posts made by each individual user on Reddit. The posts we use for our analysis come from the time period between November 2016 until March 2017. 

\subsection{RQ1: Reddit Engagement}
Our first research question explores different set of statistics to understand the engagement on the subreddits related to artificial intelligence. Towards this goal, we first understand whether the posts are questions or opinions. We utilize the `wh' questions along with the post that matches the '?' pattern. 52.8\% of the posts we crawled from Reddit related to AI are questions. This suggests that the interest to seek information on Reddit is high.


Each post made on the subreddit can be voted as up or down which also allows users to comment on this post. The score of a post is defined as the sum of up votes and down votes. In our dataset, {\em there are no posts which have a down vote}. This might be due to the inherent nature of subreddit that we are focusing on. On average, each post received around 20 votes with a median of 7 votes. The post made by Google Brain team encouraging the community to ask questions about machine learning received the maximum number of votes (4196). 

\subsection{RQ2: Optimistic or Pessimistic}
\begin{figure*}[ht!]
    \centering
    \begin{subfigure}[t]{0.3\textwidth}
        \centering
        \includegraphics[height=1.2in]{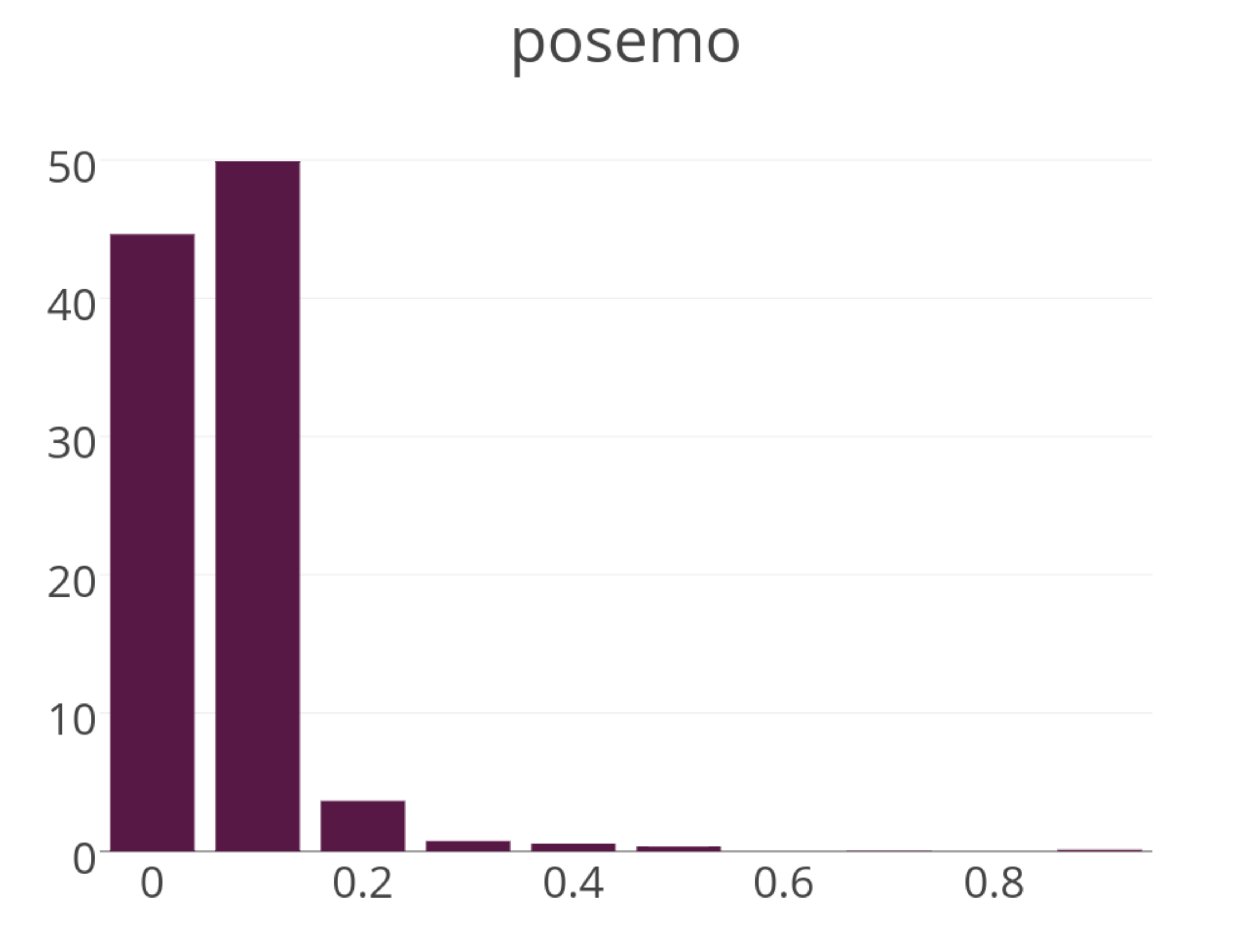}
	\caption{Positive Emotions}
    \end{subfigure}
    ~ 
    \begin{subfigure}[t]{0.3\textwidth}
        \centering
        \includegraphics[height=1.2in]{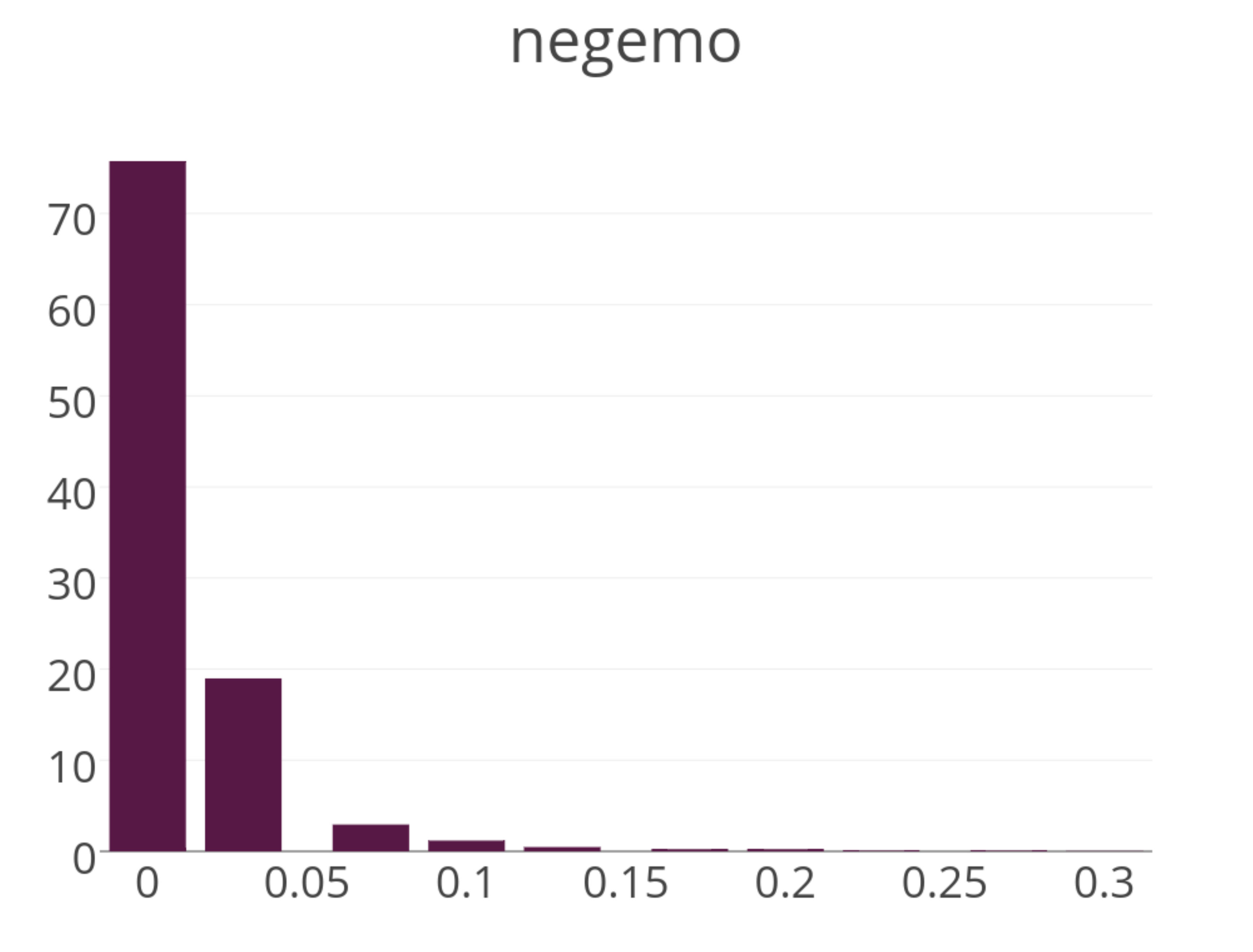}
	\caption{Negative Emotions}
    \end{subfigure}
    ~ 
    \begin{subfigure}[t]{0.3\textwidth}
        \centering
        \includegraphics[height=1.2in]{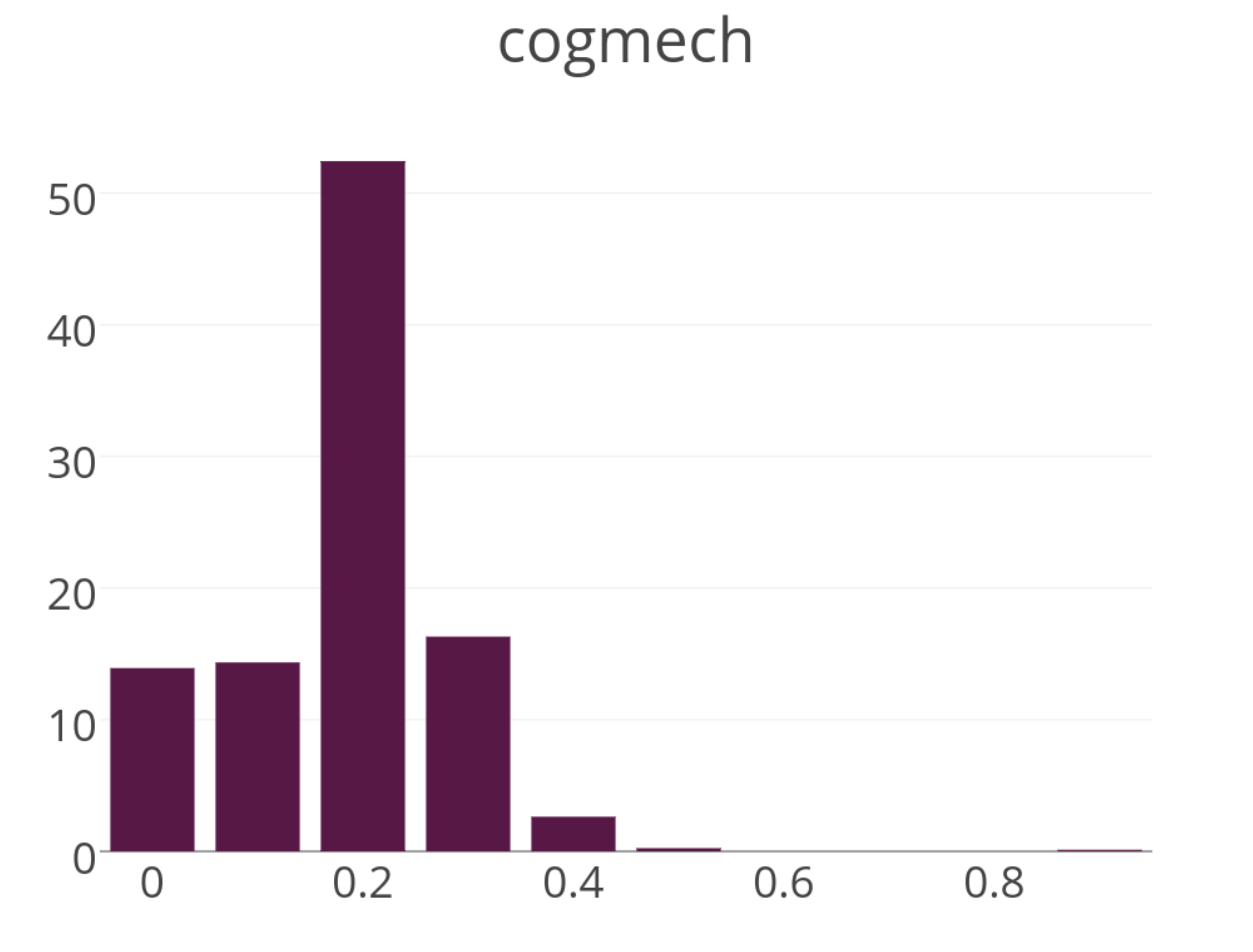}
	\caption{Cognitive Mechanisms}
    \end{subfigure}
    \caption{10 statistically significant emotions and their values measured in the posts made on Reddit}
    \label{fig:liwcreddit}
\end{figure*}

We conduct a similar analysis on Reddit posts, that we have conducted on tweets, to measure the emotional gravity. To do this, we first pre-process the data by considering only the textual content attached to the post that doesn't consider the title of the post. Similar to the Twitter analysis, we do not remove any stopwords because in the LIWC~\cite{TausczikY2010} analysis, stopwords may count towards providing emotional insights. Figure~\ref{fig:liwcreddit} shows that, Reddit posts are emotionally more positive than Twitter. 

\subsection{RQ3: Topics heavily discussed and concerning to the public}
Analysis on emotions shows that there is a higher percentage of positive emotions alongside cognitive complexity and expressing insights. To explain this phenomenon, exploring the topics focused in the posts can be beneficial. To do this, we consider the Twitter LDA~\cite{Zhao2011Traditional} package and the extracted topics are shown in Table~\ref{tab:redditvocab}. The posts about \emph{data science and careers} are the most discussed topics. We define the \emph{unclassified text} as the non-english text or a post that includes emoticons. However, we also notice that Reddit posts on Reddit focus on various topics of AI -- \emph{comparison of intelligent systems with humans} (Topic-6), \emph{properties of intelligent machines} (Topic-7), \emph{how to learn or train these intelligent models} (Topic-8), \emph{neural networks} (Topic-4), \emph{applications like games} (Topic-3), \emph{how industry is spreading its interest around AI} (Topic-5), \emph{future applications} that could be envisioned (Topic-2).

\begin{table*}[ht!]
\centering
\small
\begin{tabular}{ l | l |  p{7.4cm} | l }  
\textbf{ID} & \textbf{Topic} & \textbf{Top Tags} & \textbf{\% of tweets} \\ \hline
1 & Data science careers & data, math, science, degree, python, statistics, school, phd, stats, scientist & 23.3\% \\ \hline
2 & Future intelligence & human, intelligence, future, neural, brain, create, learning, talking, years, quantum & 11.5\% \\ \hline
3 & Games & people, work, code, game, learns, find, create, play, brain, function & 10.2\% \\ \hline
4 & Neural networks \& deep learning & training, network, model, neural, function, layer, train, gradient, batch, weights, images & 9.5\% \\ \hline
5 & Industry \& AI & google, system, alphago, brain, world, data, neural, state, memory, tensorflow & 9.3\% \\ \hline
6 & Humans \& intelligent machine & human, brain, consciousness, people, intelligence, model, computer, neural, utility, information & 9.2\% \\ \hline
7 & Intelligent machines & humans, system reasoning, algorithm, information, agents, smart, control, turing, robot, intelligence & 6.1\% \\ \hline
8 & Learning \& training & model, attributes, data, features, regression, layers, forest, network, function, error & 6.0\% \\ \hline
9 & Unclassified & unclassified & 4.6\% \\ \hline
\end{tabular}
\caption{9 topics and the corresponding vocabulary extracted along with the percentage distribution of reddit posts across these topics.}
\label{tab:redditvocab}
\end{table*}

In comparison to Twitter that focuses only on 7.2\% of tweets on {\em data science}, Reddit has higher percentage of posts on this topic. However, 7 of the 9 topics (that constitute 72.1\% of the Reddit posts) very tightly focus on the technical details about intelligent systems and their performance. For example -- details about training the learning models, properties of intelligence machines, nuances of technicalities from the perspective of industry, emphasizes on technical aspects of AI systems. On the other hand, Twitter topics are relatively skewed in terms of mostly focusing on marketing, recruitment, stories, impact, etc and do not heavily focus on the technical details about the intelligent systems. This might be due to the fact that Reddit as a platform does not have any constraints on the length of the post. This could be one of the reasons why the posts on Reddit focus on in-depth technicalities about the AI systems.

\subsection{RQ4: User Analysis}
As mentioned earlier, understanding the conclusions we inferred from the Reddit posts could be more valuable if we recognize the interests of users. To understand the background of the 5,382 unique set of Reddit users from our dataset, we crawl their recent 100 Reddit posts.

\begin{enumerate}
\item {\em \textbf{Topic-1} [Topics about life]: people, good, pretty, money, games, things, years, lot, high, pay
\item \textbf{Topic-2} [Political issues]: government, country, world, time, years, women, good, bad, person, life
\item \textbf{Topic-3} [Mobile or Web Applications]: work, love, great, windows, app, phone, video, version, link, find
\item \textbf{Topic-4} [Training neural networks]: model, code, neural, function, training, network, deep, learn, image
\item \textbf{Topic-5} [Intelligence aspects]: ai, people, human, brain, work, learning, data, intelligence, understand
\item \textbf{Topic-6} [Jobs and programs offered in Data Science]: science, python, job, math, experience, programming, courses, statistics, degree, ml
\item \textbf{Topic-7} [External URL shares]: en.wikipedia.org, www.youtube.com, watch, i.imgur.com, delete, youtu.be, definition}
\end{enumerate}

\begin{figure}[ht!]
\centering
\includegraphics[height=1.2in]{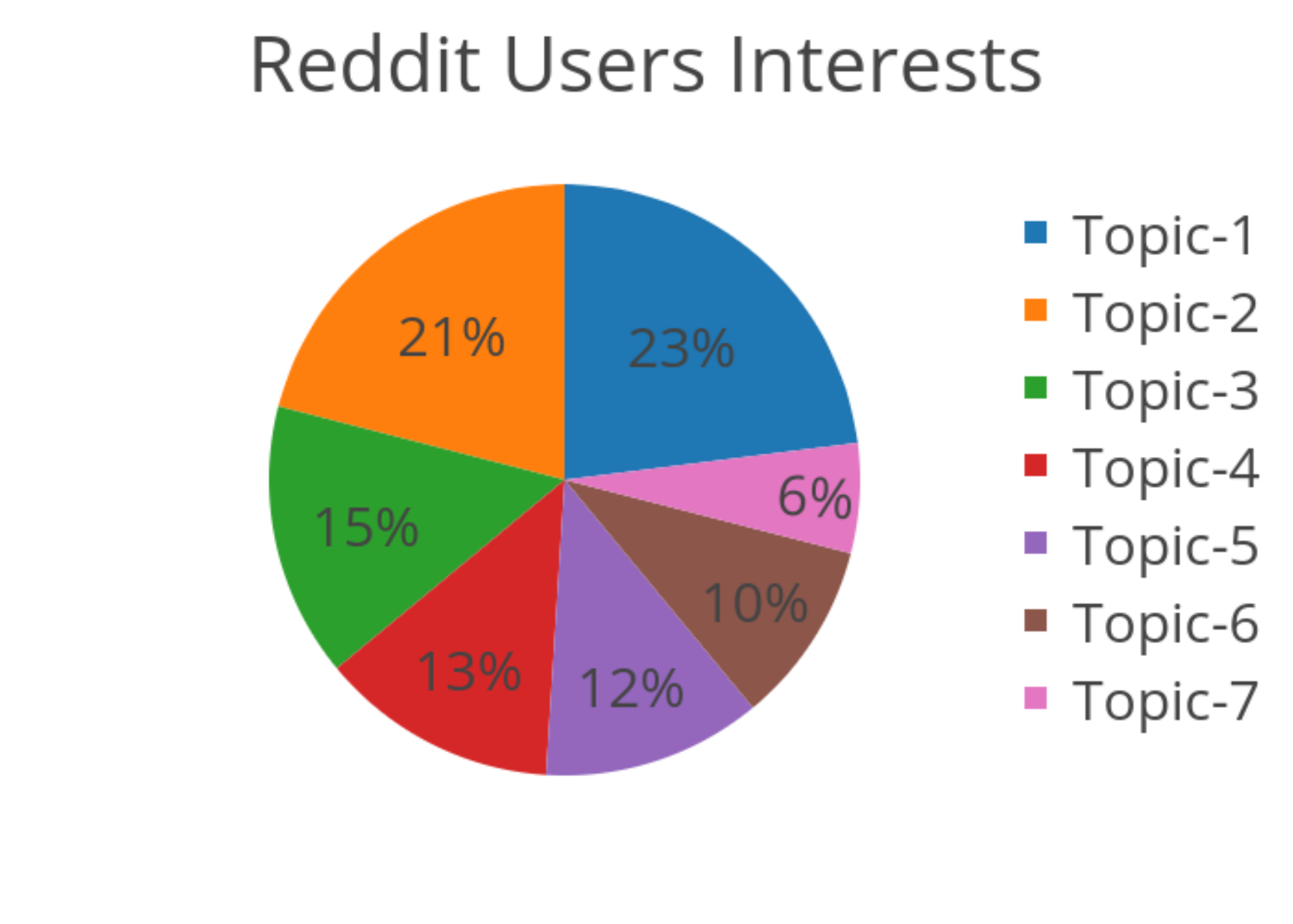}
\caption{Topic Distributions Extracted from Reddit. Topic-1: Topics about life ; Topic-2: Political issues ; Topic-3: Applications ; Topic-4: Training neural networks; Topic-5: Intelligence aspects; Topic-6: Datascience related Jobs and programs ; Topic-7:External url shares }
\label{fig:reddtop}
\end{figure}

According to the topic distributions shown in Figure~\ref{fig:reddtop}, these users are interested in technology -- especially neural networks. Eventhough on an average, 44\% of the reddit posts made by a user focuses on general topics about \emph{life and politics}, 25\% of their posts talk about \emph{training neural networks and intelligence}. This may shed more light on the perceptions of AI that are mainly shared and discussed by users with interests in technology especially related to AI. 

\vspace{2in}
\noindent
The main findings from the Reddit analysis are that the users are more optimistic than the users on Twitter. The topic analysis shows the technical depth in to different aspects of AI. This might be due to the nature of Reddit platform that doesn't have restrictions on the length of a post. 
\end{document}